\documentclass{article}

\newcommand{\system}{\textsc{Garfield}}
\newcommand{\papertitle}{\system{}: System Support for\\ Byzantine Machine Learning}

\newcommand{\mkrum}{\textit{Multi--Krum}}
\newcommand{\krum}{\textit{Krum}}
\newcommand{\kardam}{\textit{Kardam}}
\newcommand{\bulyan}{\textit{Bulyan}}
\newcommand{\medianoid}{\textit{Median}}
\newcommand{\brute}{\textit{MDA}}

\newcommand{\aggregathor}{\textit{AggregaThor}}
\newcommand{\byzsgd}{\textit{MSMW}}
\newcommand{\learn}{\textit{decentralized learning}}

\newcommand{\paragraphspaceheight}{0.2cm}
\newcommand{\paragraphspace}{\vspace{\paragraphspaceheight}}
\newcommand{\mathsep}{,~}

\newcommand{\setr}{\mathbb{R}}
\newcommand{\setn}{\mathbb{N}}
\newcommand{\expect}{\mathop{{}\mathbb{E}}}

\newcommand{\norm}[1]{\left\lVert{#1}\right\rVert}

\newcommand{\gradloss}{\nabla{}L}
\newcommand{\realgrad}[1]{\gradloss{}\left({#1}\right)}

\newcommand{\indexvar}[3]{{#3}^{\ifthenelse{\equal{#1}{}}{}{\left({#1}\right)}}_{#2}}
\newcommand{\range}[2]{\left[{#1} \,..\, {#2}\right]}

\newcommand{\params}[2]{\indexvar{#1}{#2}{\theta}}
\newcommand{\gradvar}[2]{\indexvar{#1}{#2}{g}}

\newcommand{\marker}[2]{\paragraphspace\par\noindent\fbox{\parbox{0.95\columnwidth}{\textsc{{#1}:} \textit{#2}}}\paragraphspace\par}
\newcommand{\todo}[1]{\marker{ToDo}{#1}}

\usepackage{filecontents}
\usepackage{graphicx}
\usepackage{balance}
\usepackage{diagbox}
\usepackage{url}
\usepackage[utf8]{inputenc}
\usepackage[english]{babel}
\usepackage{xr,xfrac,xspace,mathtools,amsmath,amsfonts,amssymb,bm}
\usepackage{amsmath}
\usepackage{amsthm}
\usepackage{thmtools, thm-restate}
\usepackage{amssymb}
\usepackage{xspace}
\usepackage{booktabs}
\usepackage{epsfig}
\usepackage[normalem]{ulem}
\usepackage[font=small]{subfig}
\usepackage[font=bf]{caption}
\usepackage{float}
\usepackage[group-separator={,}]{siunitx}
\usepackage{xcolor,colortbl}
\usepackage{lipsum}
\usepackage{capt-of}
\usepackage{algorithm}
\usepackage{listings}
\usepackage{multicol}
\usepackage{multirow}
\usepackage[normalem]{ulem}
\usepackage{soul}
\usepackage{relsize}
\usepackage{stmaryrd}
\usepackage{wrapfig}
\usepackage{array}
\usepackage{enumitem}
\usepackage{verbatim}
\usepackage{authblk}
\providecommand{\keywords}[1]
{
  \small	
  \textbf{\textit{Keywords---}} #1
}

\title{\papertitle{}}

\author{Rachid Guerraoui, Arsany Guirguis\thanks{Corresponding author.}, J\'{e}r\'{e}my Max Plassmann, Anton Alexandre Ragot, S\'{e}bastien Rouault}
\affil{{EPFL} \\ firstname.lastname@epfl.ch}
\date{}

\DeclareMathOperator*{\argmin}{arg\,min}

\begin{document}
\newtheorem{theorem}{Theorem}[section]
\newtheorem{lemma}[theorem]{Lemma}

 \lstset{language=Python,
    frame=lines,
    basicstyle=\ttfamily\small,
    numbers=left,
    keywordstyle=\color{blue},
    commentstyle=\color{gray},
    stringstyle=\color{red},
    showstringspaces=false,
    identifierstyle=\color{black},
    numbersep=0.8em,
    xleftmargin=1.5em}
    
\maketitle
\begin{abstract}

We present \system{}, a library to transparently make 
machine learning (ML) applications, 
initially built with 
popular (but fragile) frameworks, e.g.,\ 
TensorFlow and PyTorch, Byzantine--resilient. 
\system{} relies on 
a novel object--oriented design, 
reducing the coding effort, and
addressing the vulnerability of the shared--graph architecture followed by classical ML frameworks.
\system{} encompasses various communication patterns 
and supports computations on CPUs and GPUs,
allowing addressing
the general question of the 
practical cost of Byzantine resilience in 
ML applications. 

We report on the usage of \system{} on three main ML architectures:
(a)~a single server with multiple workers, 
(b)~several servers and workers, 
and 
(c)~peer--to--peer settings. 
Using \system{}, we highlight 
interesting facts about the cost of Byzantine resilience.
In particular, (a)~Byzantine resilience, unlike crash resilience, induces an accuracy loss, 
(b)~the throughput overhead comes 
more from communication 
than from robust aggregation, and
(c)~tolerating Byzantine servers 
costs more
than tolerating Byzantine workers.
\end{abstract}

\section{Introduction}
Machine Learning (ML) is nowadays distributed~\cite{li2014scaling,meng2016mllib}. A major motivation is scalability. The quantity of data available to ML tasks is huge and can only be handled with distributed architectures. For instance, the size of Google's ad impression log to train an ad click predictor could reach trillions of  examples~\cite{kim1many}, each representing a high-dimensional feature vector.
Such a dataset  expands daily with new examples~\cite{mcmahan2013ad} in order to yield better models. 

A now classical approach to distribute an ML task is through the \emph{server/worker} architecture~\cite{li2013parameter}. A parameter server coordinates the distribution of the training task among a large set of worker nodes. The parameter server typically aggregates the workers' \emph{gradients} by merely averaging them~\cite{abadi2016tensorflow,bonawitz2017practical}, following the standard workhorse optimization algorithm: Stochastic Gradient Descent (SGD)~\cite{rumelhart1986learning}. 
An alternative, more decentralized, approach does not distinguish 
servers and workers. Each node has a copy of the model and keeps its data locally~\cite{patarasuk2009bandwidth,vanhaesebrouck2016decentralized}, typically to protect it and save bandwidth or devices' batteries. 
In this approach, all nodes apply SGD and communicate (in a peer--to--peer fashion) what they learned so far to refine their models,\footnote{Also called \emph{collaborative learning}~\cite{colalearning}.} usually also through averaging.



As the number of participating machines in a distributed setup increases, so does the probability of failure of any of these machines. In distributed computing, the most general way to model such failures is to assume an adversary that can control a subset of the system and make it arbitrarily deviate from the normal execution: we talk about Byzantine failures~\cite{lamport1982Byzantine}. This includes bogus software, faulty hardware as well as malicious attacks. With the increasing use of ML in mission-critical applications~\cite{esteva2017dermatologist,bloom2017self,rao2018deep}, building robust systems against these kinds of failures becomes a necessity.
Using vanilla \emph{state machine replication} (SMR) to solve such a problem was shown impractical in the ML context~\cite{aggregathor}.

Tolerating Byzantine workers without replicating them has been recently well studied in the convex (see e.g.,\ \cite{chen2017distributed,alistarh2018byzantine}) as well as non-convex (see e.g.,\ \cite{krum, rajput2019detox, lilisu2019,xie2019zeno++}) settings. 
A key idea is to replace the vulnerable averaging scheme, 
to aggregate gradients, 
by a \emph{statistically robust gradient aggregation rule} (GAR),
e.g.,\ \medianoid{}~\cite{xie2018generalized}.
Several GARs were proposed, and they differ according to their computation cost, 
the assumptions they make on the dimension of the model, 
or the ratio of correct nodes~\cite{baruch2019little,xie2019fall,bulyanPaper}. 
The idea was then extended to tolerate Byzantine servers, e.g.,\ in~\cite{el2020genuinely}, as well as to the decentralized settings with no distinction between servers and workers, e.g.,\ ~\cite{yang2019bridge,yang2019byrdie,el2020collaborative}.

Most work on Byzantine resilient ML has however been theoretical and, it is not clear how to put the published 
algorithms to work, especially in the pragmatic form of library extensions to existing, and now classical, ML frameworks, namely TensorFlow~\cite{abadi2016tensorflow} and PyTorch~\cite{paszke2019pytorch}. 
These frameworks share two specific characteristics that go against Byzantine resilience.
First, and for performance reasons, they rely on a shared memory design. For instance, TensorFlow uses one shared computation graph among all machines to define the learning pipeline. 
Such a design is problematic as Byzantine nodes can corrupt the learning state at honest ones. 
Second, most of the high--level communication abstractions given by such frameworks assume trusted, highly--available machines.
For instance, the distributed library of PyTorch allows for collective communication among processes, yet such calls block indefinitely in case of a process crash or network failure.

We present in this paper \system{}, a library that enables the development of Byzantine ML applications on top of popular frameworks such as TensorFlow and PyTorch, while achieving \emph{transparency}: applications developed with either framework do not need to change their interfaces to tolerate Byzantine failures.
\system{} relies on an object--oriented design, which makes it possible 
to adapt the components in the network, and how they interact,
according to the desired Byzantine resilient scheme. In particular, 
\system{} allows for both synchronous and asynchronous communication.
\system{} also includes several \emph{statistically-robust gradient aggregation rules} (GARs),\footnote{Hence the name: GARfield.} which can be used to 
filter out the effect of Byzantine replies.
\system{} also supports full--stack computations on both CPUs and GPUs. 
Along our \system{} implementation journey, we took several design decisions to promote its practicality.
For instance, we implemented specific schemes to parallelize Byzantine--resilient GARs, 
especially on GPUs. 
Moreover, we devised the notion of \emph{separate replicated graphs} for TensorFlow rather than relying on its \emph{shared graph} design, as the latter would be a killer in a fully Byzantine environment, i.e.,\ without any trusted machine~\cite{aggregathor}.
We rely on gRPC for point--to--point pull--based communication, due to its speed and because it is currently the defacto standard communication method for popular ML frameworks.
Our implementation parallelizes RPC calls  
to improve the scalability of algorithms implemented with \system{}.
Finally, we carefully manage CPU memory to minimize memory copying and allow for faster algorithms. 




We report on the usage of \system{} on three ML architectures: 
(1) 
tolerating Byzantine workers while assuming one trusted, central server, 
(2) replicating the parameter server to account for Byzantine servers as well as Byzantine workers, and 
(3) considering a peer-to-peer, decentralized setting with no distinction between 
servers and workers. 

We  report on our evaluation of \system{}, addressing the general question of the very practical cost of Byzantine resilience in a distributed ML deployment when compared to a vanilla deployment where all components are trusted. 
We consider various ML models, 
as well as different hardware, i.e.,\ CPUs and GPUs. We also study the cost of different degrees of resilience. 


Essentially, we show that Byzantine resilience introduces up to $10\%$ loss in accuracy compared to non--Byzantine deployments. In contrast, crash resilience does not introduce any such loss. 
As we  show in the paper, such an observation is  obvious only with large--scale deployments.
In terms of throughput, we quantify the overhead of various Byzantine resilience degrees, compared to a vanilla deployment. 
Basically, we find that tolerating Byzantine servers induces much more overhead than tolerating Byzantine workers.
For instance, we quantify the cost of adding Byzantine resilience to servers, compared to tolerating only Byzantine workers with a trusted server, to $53\%$, and the cost of Byzantine resilience, compared to the crash--tolerant baseline, to $22\%$ (using GPUs). 
We root the resilience overhead mainly to communication.
Concretely, our experiments show that communication accounts for more than $75\%$ of the overhead while robust aggregation contributes to only $11\%$ of such an overhead. 
We also highlight the very fact that Byzantine algorithms in a peer-to-peer setup do not scale, unlike those following the parameter server architecture.
We also report on the fact the overhead of Byzantine resilience depends more on the number of participating nodes, be they workers or servers,  than on the model dimension.  
Notably, using GPUs achieves a performance improvement of at least one order of magnitude over CPUs.

\paragraphspace
\textbf{\emph{Summary of contributions.}}
\newline
\noindent \textbf{1.} We introduce \system{}, a library to build Byzantine--resilient ML applications, following a novel object--oriented design and introducing flexible communication abstractions. 

\noindent \textbf{2.} We integrate \system{} with 
TensorFlow and PyTorch, while achieving (a) transparency: ML applications need not change their interfaces to work with \system{} and (b) full-stack computation support on CPUs and GPUs.

\noindent \textbf{3.} We evaluate our implementation of \system{} 
using various models, baselines, and hardware infrastructure, analyzing the cost of Byzantine resilience in different scenarios. The code of \system{} is available at~\cite{code}. 

\section{Background}
\label{sec:background}
\subsection{Stochastic Gradient Descent}
\label{sec:background-sgd}

Stochastic Gradient Descent (SGD)~\cite{rumelhart1986learning} is the most widely-used \emph{optimization} algorithm in ML applications~\cite{chilimbi2014project,li2014scaling}.
It becomes the defacto standard to optimize objective functions that can be used with ML techniques such as neural networks.

To explain how SGD works, assume the objective function (also called \emph{loss} function) to be
$L\left( \bm{x} \right) \in \setr$. This basically measures ``how incorrect the model 
is when labeling an input''.
SGD addresses the following optimization problem:
\begin{equation}
\label{eq:opt}
\bm{x}_{opt} \triangleq \argmin_{\bm{x} \in \mathbb{R}^d} L(\bm{x})
\end{equation}
The procedure is iterative: in each step $k$, one can 
    estimate the gradient $G\left(\bm{x^{(k)}}, \xi\right)$, 
    with a subset $\xi$ of size $b$ of the training set, called \emph{mini--batch}. This represents an approximation of the \emph{uncomputable} real gradient.  
    Then, using the estimated gradient, the model parameters ($\bm{x}$) are updated as follows:
    \begin{equation}
        \label{eq:opt-step}
        \bm{x}^{(k+1)} = \bm{x}^{(k)} - \gamma_k \cdot G\left( \bm{x}^{(k)}, \xi \right),
    \end{equation}
    where $\{\gamma_k\}$ is called the \emph{learning rate}.

\subsection{Distributed Machine Learning}

Estimating the gradient at $\bm{x}$ is computationally expensive, given the big datasets and the complex high--dimensional models we have nowadays.
Fortunately, this gradient estimation is easily parallelizable.
Basically, $n$ machines can each partially estimate the gradient using a mini--batch of size $\sfrac{b}{n}$, which can be then aggregated together to restore the complete estimation.

One widely used architecture enabling this distribution is the \emph{parameter server} scheme~\cite{li2013parameter}, 
where a centralized server holds the parameters~$\bm{x}$, and the other machines (called workers) own data batches.
For each training step, the server first broadcasts the parameters to workers,
which then share the heavy gradient estimation.  
When a worker completes its estimation, it sends it to the parameter server, which
finally aggregates the received estimations (typically by \emph{averaging}) and updates the parameters $\bm{x}$, as in Equation~(\ref{eq:opt-step}).

Another variant to distribute gradient estimation is the decentralized learning~\cite{vanhaesebrouck2016decentralized} in which all machines collaborate together to train a model without a central entity.
In this scheme, each machine owns a copy of the model and some local data that is never shared with the others.
In each training step, each machine estimates a gradient (using SGD) and shares it, in a peer--to--peer fashion, with the others.
Gradient estimations are then aggregated locally on each machine and are used to update the parameters $\bm{x}$.


\subsection{Byzantine Resilience}
In the parlance of classical distributed computing, a system tolerates a Byzantine fault when it copes with a machine that can deviate arbitrarily from the algorithm assigned to it~\cite{lamport1982Byzantine}.
Such a behavior abstracts any kind of failures, including software bugs, hardware defects, corrupted data~\cite{ghiassi2019robust}, communication omissions, or even adversarial attacks. 
We consider the ML context where any of the machines contributing to the learning process 
can behave arbitrarily.
In such context, tolerating Byzantine behavior basically ensures following the same learning path that would have been achieved in the absence of Byzantine machines.
Previous work achieved that using robust aggregation, e.g.,~\cite{krum,li2019rsa}, redundant gradient computation along with coding schemes~\cite{chen2018draco}, combining both ideas~\cite{rajput2019detox}, or performance--based ranking~\cite{xie2018zeno,xie2019zeno++}.
\section{The \system{} Design}
\system{} is a library to build Byzantine--resilient SGD--based ML applications that rely on (1) robust aggregation and (2) communication of plain vectors (without compression nor quantization). 
We first introduce a few \emph{statistically--robust gradient aggregation rules (GARs)} that are included in \system{} and then, we discuss the modular design of \system{}.
\subsection{Statistically Robust GARs}
\label{subsec:gars}
A GAR is merely a function of $\left( \setr^d \right)^q \rightarrow{} \setr^d$, with $d$, the dimension of the input vector space $\setr^d$ (i.e.,\ a gradient or a model), and $q$, the number of input vectors to be aggregated. 
Basically, these GARs wait for $q$ vectors before applying some functions on them.
Hence, in synchronous, non--faulty settings, these GARs can be deployed with $q$ machines in the system (so that the aggregator node can gather replies from \emph{all} nodes in the system within some time bound).
Yet, in asynchronous settings, one would require to deploy $q+f$ nodes to use these GARs, to ensure liveness of the protocol, where $f$ denotes the maximum number of Byzantine inputs.
In short, all GARs output a vector with special statistical properties that make them safe to use in the Byzantine setting.

\noindent{\textbf{1. \medianoid{}~\cite{xie2018generalized}}} computes the coordinate-wise median among the input gradients and outputs one gradient of these medians. 
\medianoid{} requires $q \ge 2f+1$, and its asymptotic complexity is $\mathcal{O}(qd)$.

\noindent{\textbf{2. \krum{}~\cite{krum}}} assigns a \emph{score} to each gradient (based on a sum of distances with the closest neighbors), 
and then returns the smallest scoring gradient. 
\mkrum{} (a variant of \krum{}) achieves a better convergence rate than \medianoid{}~\cite{aggregathor}.
It requires however $q \ge 2f+3$, and its asymptotic complexity is $\mathcal{O}(q^2d)$.

\noindent{\textbf{3. \brute{}~\cite{rousseeuw1985multivariate}}} finds a subset group of gradients of size $q - f$ with the minimum diameter among all other subsets, where the diameter of a group is defined as the maximum distance between any two gradients of this subset. \brute{} then outputs the average of the chosen subset.
Notably, \brute{} carries an exponential\footnote{Exponential when $f = \mathcal{O}\!\left( q \right)$, polynomial when $f = \mathcal{O}\!\left( 1 \right)$.} asymptotic complexity of $\mathcal{O}\!\left( \binom{q}{f} + q^2 d \right)$.
Yet, as we will discuss later, its assumptions about variance are weaker than for the previous two GARs.
It requires $q \ge 2f+1$.

\noindent{\textbf{4. \bulyan{}~\cite{bulyanPaper}}} robustly aggregates $q$ gradients by iterating several (say $k$) times over another Byzantine--resilient GAR, e.g.,\ \mkrum{}.
In each of these $k$ iterations, \bulyan{} extracts the gradients selected by such a GAR. 
Then, it computes the coordinate-wise median of the $k$ selected gradients.
It then extracts the closest $k'$ gradients to the computed median, and finally returns the coordinate-wise average of these $k'$ gradients.
Unlike previous GARs, 
\bulyan{} can sustain a model with a large dimension.
Yet, it requires $q \ge 4f+3$, and its asymptotic complexity is $\mathcal{O}(q^2d)$.

\noindent{\textbf{Tradeoffs.}}
In addition to the differences in the ratio of tolerated Byzantine nodes 
(inequalities relating $q$ with $f$)
and the computational cost of each GAR, the model dimension is also crucial in deciding which GAR to use.
For high dimensions (e.g.,\ order of millions) and a strong adversary, one should use \bulyan{}.
In low dimensions, the application setup should satisfy the \emph{variance} assumption of the working GAR, as given below:
\begin{align*}
    \exists \kappa \in \left] 1, +\infty \right[ \mathsep \forall \left( i, t, \params{}{} \right) \in \range{1}{n-f} \times \setn \times \setr^d \mathsep \\
    \kappa \, \Delta \, \sqrt{\expect\left( \norm{\gradvar{i}{t} - \expect\gradvar{i}{t}}^2 \right)} \le \norm{\realgrad{\params{}{}}},
    \end{align*}

where,
\begin{equation*}
    \Delta = \begin{cases}
        \frac{2\sqrt{2} f}{n - f} &
        \text{if GAR = \brute{}} \\
        \sqrt{2 \left( n\!-\!f\!+\!\frac{f \left( n - f - 2 \right) + f^2 \left( n - f - 1 \right)}{n - 2 \, f - 2} \right)} &
        \text{if GAR = \krum{}} \\
        \sqrt{n-f} &
        \text{if GAR = \medianoid{},}
    \end{cases}
\end{equation*}
where $\gradvar{i}{t}$ is the estimated gradient by worker $i$ at time $t$, and $L(\theta)$ is the loss function at the model state $\theta$.
We provide a simple tool, \texttt{measure\_variance.py}, to estimate whether such a condition is satisfied.
Such a tool takes the experimental setup ($n$, $f$, the batch size, ..etc) as inputs. 
Then, it does few training steps, while estimating the true gradient $\realgrad{\params{}{}}$ by computing a gradient using a huge batch size.
In each training step, such a tool checks whether the condition stated above is satisfied or not, and then, gives the user some statistics on how often such a condition is satisfied with each GAR.

\begin{figure}[t]
\centering
\vspace{-2mm}
\includegraphics[width=0.45\textwidth]{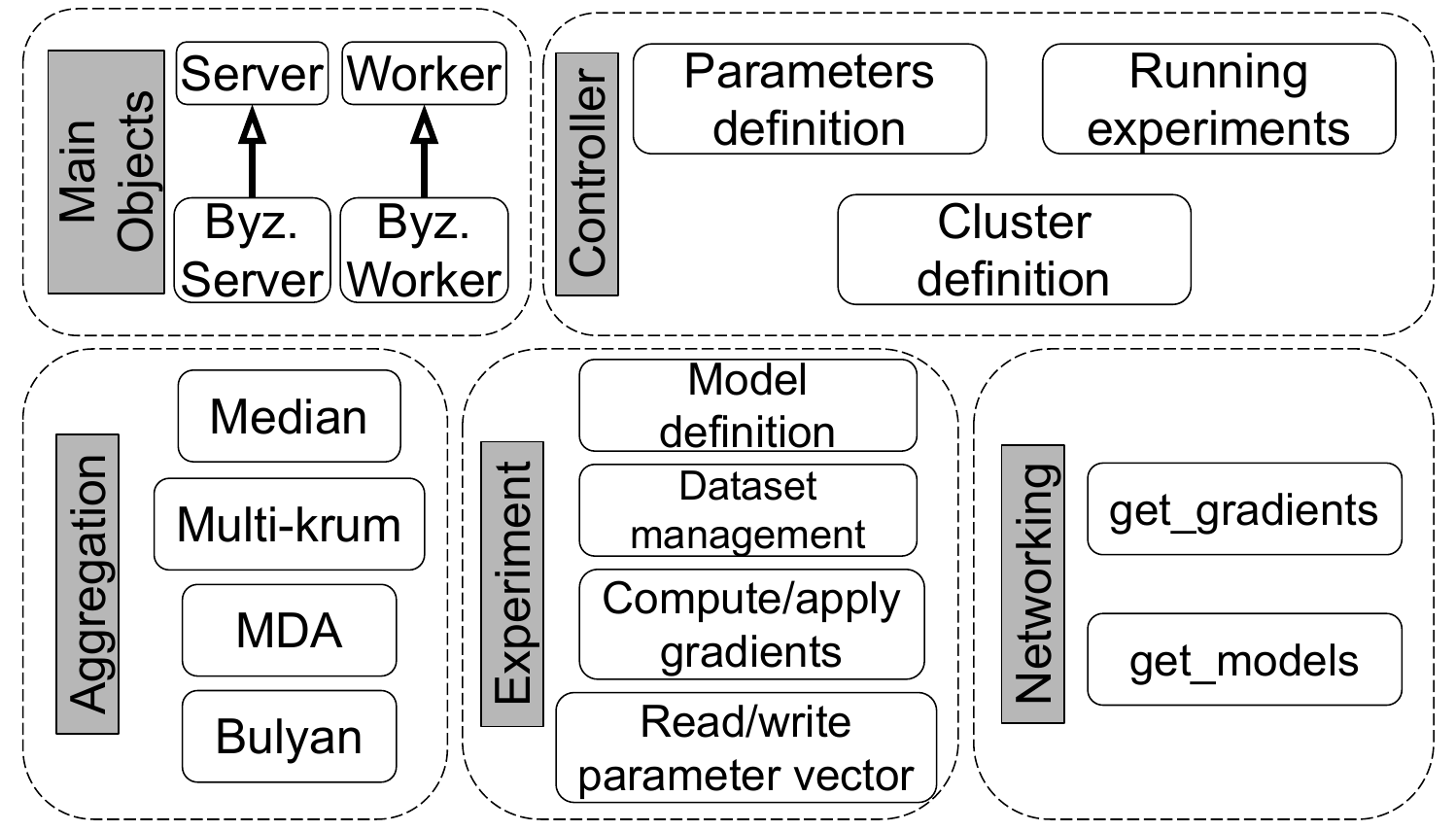}
\caption{The \system{} system components.}
\label{fig:arch}
\vspace{-4mm}
\end{figure}

\subsection{System Components}
\system{} is designed in a modular way as shown in Figure~\ref{fig:arch}.
In this section, we describe each of these modules.


\paragraph{Main objects}
\system{} defines two main objects that can be used for learning: Server and Worker.
The server is responsible for storing and updating the model state while workers train this model using their own local data.
The server typically initiates a learning step by asking a few workers to compute a gradient estimate given the model state it owns.
For this, the server object exposes the method \emph{get\_gradients()}, which we describe later in detail.
Some algorithms require servers to exchange their models states, and hence 
the server object comes with the method \emph{get\_models()}, which is the second key method in our \emph{Networking} interface, as we describe below.
In addition to the \emph{Networking} methods, the server object exposes methods to (1) update the model state, given some optimizer and a gradient estimate, (2) re-write the model, which is useful in the case of multiple server replicas, and (3) compute accuracy, given the model state and a test set.

The worker design on the other hand is much simpler. 
The worker object is passive in the sense that it only responds to requests of the server.
Basically, the worker owns some data and defines some loss function.
Its main job is to compute a gradient estimate, when asked by the server, using the data chunk it owns.
The worker then replies to the server by the gradient estimate it computed.

To support experimenting with Byzantine behavior, \system{} defines two objects: \emph{Byzantine Server} and \emph{Byzantine Worker}, which inherit from our main objects, Server and Worker respectively.
Both objects implement the popular attacks published in the Byzantine ML literature including simple ones like reversing vectors, dropping vectors, or random vectors~\cite{aggregathor,el2020genuinely} and the state--of--the--art attacks like \emph{little is enough}~\cite{baruch2019little} and \emph{fall of empires}~\cite{xie2019fall}.

\paragraph{Networking}
Existing networking abstractions in both frameworks, TensorFlow and PyTorch, are not enough to be used (1) in a Byzantine, asynchronous environment and (2) with replicated parameter servers.\footnote{Not to be confused with replicated graphs in TensorFlow. In some cases, the server needs to be replicated where these replicas (all have the same graph) are independent, rather than shared between machines, and they do exactly the same computation on the same data for fault tolerance rather than for performance. Both kinds of replication can be combined, but this is out of the scope of this paper.} 
For example, one can deploy distributed training on PyTorch using \emph{DistributedDataParallel()} or on TensorFlow by running \emph{ParameterServerStrategy()}. However, both are high--level abstractions that assume trusted, always--available machines.

\system{} (as a part of the Server class) supports two abstractions to handle communication: \emph{get\_gradients()} and \emph{get\_models()}. 
The first one is used 
to read the computed gradients by the workers.
It accepts two parameters: $t$, the index of the current iteration, and $q_w$, the number of workers from which a server should receive replies with $q_w \le n_w$ ($n_w$ denotes the total number of workers); $q_w = n_w$ denotes synchronous communication with no faults in the system, i.e.,\ a server is expecting to receive replies from \emph{all} workers. 
This function then returns the fastest $q_w$ gradients it receives.
The second abstraction works in the exact same way, yet fetching models from servers instead of gradients from workers. 
Both abstractions then enable easy and natural communication among all machines in the network in both synchronous and asynchronous settings (as we describe in Section~\ref{sec:applications}).
We give the implementation details of these abstractions in Section~\ref{sec:impl}. 

\paragraph{Aggregation}
\system{} implements the four Byzantine--resilient GARs mentioned above, on both CPUs and GPUs. We create wrappers (including dependency management, automatic compilation and loading) to use them as \emph{custom operations} 
in both TensorFlow and PyTorch.
Such wrappers make it possible to involve the GARs with the same interface for both frameworks, though the lower--level interfaces each framework provides differ substantially.

To use a GAR, the common interface consists in two functions: \emph{init()} and \emph{aggregate()}.
The \emph{init()} function takes the name of the required GAR (e.g., ``median''), the value of $n$, the total number of inputs, and $f$, the maximum number of Byzantine inputs. 
The second function, \emph{aggregate()}, takes $n$ \emph{tensors} (could represent gradients or models) and outputs the aggregated one.
Whether this function will execute on a CPU or a GPU depends on the device on which the input vectors are stored.
In this way, our design abstracts the device, CPU or GPU, and the framework, TensorFlow or PyTorch, away from the developer. 

\paragraph{Controller}
This module controls the training task. It is used for cluster deployment, definition of parameters, as well as for launching experiments.
This encompasses parsing the cluster information, such as nodes' jobs (servers or workers) as well as their IPs and port numbers, starting the training procedure over SSH, and parsing experiments' parameters, e.g.,\ the maximum number of Byzantine workers and servers. 

\paragraph{Experiment}
This module abstracts the available models and datasets for training. Our design gives a unified interface to the TensorFlow \emph{slim} research models, \emph{Keras}, and \emph{TorchVision} models. This enables us to experiment with various models, e.g.,\ 
ResNet~\cite{DBLP:journals/corr/HeZR016} and VGG~\cite{simonyan2014very}. 
On the training side, we leverage the \emph{compute} and \emph{apply} gradients functions to the underlying system, be it TensorFlow or PyTorch.
\section{The \system{} Implementation}
\label{sec:impl}



First, we present how we implement the communication abstractions, i.e.,\ \emph{get\_models()} and \emph{get\_gradients()} in TensorFlow and PyTorch. Then, we show how we implement an efficient version of the \emph{median} function (which is used in \medianoid{} and \bulyan{} GARs) on GPUs.\footnote{We also include GPU implementation of other GARs yet, we focus on \emph{median} because its GPU implementation is challenging as we discuss.}
Finally, we discuss some tricks we employ for better memory management.
Our code is available~\cite{code} and will be open--sourced upon publication.

\subsection{Communication in TensorFlow}
TensorFlow adopts the notion of a shared dataflow graph in which all computations are defined in one graph, even if deployed in a distributed environment, where all participating nodes share this graph. This is a critical vulnerability in the Byzantine setting as Byzantine nodes can write and execute code on the other honest nodes~\cite{aggregathor}. Also, such shared graph abstraction hides the data communication among workers and servers, reducing the programming flexibility and disallowing having multiple communication rounds per learning step, which is crucial for Byzantine resilience~\cite{el2020genuinely}.

We follow another route in which all nodes create an independent, yet replicated graph. 
Though this design has high memory overhead, we believe it is necessary to tolerate adversarial behavior.\footnote{Such an overhead could be reduced if the environment is Byzantine--free.}
In addition to resolving the vulnerability, this design allows for more flexible communication patterns among the participating nodes. 
We use gRPC for communication and protocol buffers~\cite{protobuf} for serializing and deserializing data. We use the pull model for transferring data: when a node 
needs some data, it pulls this data from the other nodes by initiating multiple remote procedure calls to such nodes. Each node implements a server that serves these requests.
We define the protocol buffers which encode data exchanged between participants. 
We parallelize the replicated communication between workers and servers for requesting gradients and updated models so as to reduce the communication time as much as possible. 
However, abandoning the highly optimized TensorFlow distributed runtime and using independent graphs on each node require context switches between TensorFlow and Python runtimes (as protocol buffers currently cannot serialize tensors directly). Concretely, when a node is requested to send a gradient or a model, it serializes the requested data to a protocol buffer, exiting the TensorFlow graph/runtime. On the receiver side, a node deserializes the received bytes back to a tensor. Our experiments show that the overhead of these conversions (including memory copying) is non-negligible.


\subsection{Communication in PyTorch}
We implement the same abstractions in PyTorch yet with a slightly different design compared to the TensorFlow one.
First, there is no context switch between PyTorch and Python since PyTorch gives communication abstractions that can be used directly on tensors. 
Second, we pipeline the communication with aggregation (whenever possible) as PyTorch gives access to gradients of each layer in the deep network separately; this allows for better utilization of both network and computation devices and hence, better scalability. 
Third, in addition to RPC support,\footnote{PyTorch fully supports gRPC communication only in \emph{v1.6}.} our networking library also supports the distributed \emph{communication collectives} of PyTorch.\footnote{\url{https://pytorch.org/docs/stable/distributed.html}}
In the latter case, our implementation automatically chooses the best communication backend between \emph{nccl} and \emph{gloo} to allow GPU-to-GPU communication whenever possible. This is a plus compared to the RPC--based implementation as the latter does not allow communication over GPUs.

\subsection{SIMT \medianoid{} Function}

Our implementation of the \emph{median} function on CPU is quite straightforward: each of the $m \ge 1$ available cores processes a continuous share of $\sfrac{n}{m}$ coordinates.
Then each core applies, for each coordinate of its share, \emph{introselect} (or equivalent) by calling the standard C++ \texttt{std::nth\_element}.

Nevertheless, even embarrassingly parallel algorithms like \emph{median} would not necessarily benefit from running on GPGPUs (General--Purpose computing on Graphics Processing Unit).
That is because modern GPGPUs, to achieve parallel execution on many threads while limiting instruction fetch costs, batch threads into groups of, e.g.,\ 32 threads that execute the same instruction.\footnote{In case of branching, the threads execute in \emph{lock--step}.}
Algorithms like \emph{introselect}~\cite{musser1997introspective} are branch--intensive, with possibly many instructions executed in each branch, and so, fails to scale on GPUs.

Reminiscent of~\cite{branchless-median}, our implementation of \emph{median} is built around a primitive that orders 3 elements without branching, by the use of the \emph{selection instruction}, which converts a predicate to an integer value.
Let \texttt{v} be the table of size $3$ to reorder by increasing values.
Thanks to the selection instruction, we can compute \texttt{c[3] = \{v[0] > v[1], v[0] > v[2], v[1] > v[2]\}},
where \texttt{a > b} is $1$ if $a > b$ else $0$.
Then the indices \texttt{i[2] = \{ \\\hspace*{0.5cm}(1+c[0]+2*c[1]+c[2]-(c[1]$\oplus{}$c[2]))/2,\\\hspace*{0.5cm}(4-c[0]-2*c[1]-c[2]+(c[0]$\oplus{}$c[1]))/2 \}}, \\ and finally the reordered values \texttt{w} of \texttt{v} is: \texttt{w[3] = \{ \\\hspace*{0.5cm}v[i[0]], v[3-i[0]-i[1]], v[i[1]] \}}. \\
Using this reordering primitive, we manage to implement an efficient version of \emph{median} with minimal branching.

\subsection{Memory Management}
We describe here a few tricks that we use to minimize memory footprint and reduce copying data among the CPU and the GPU memory.
First, whenever possible, we pin training data to memory. 
This impacts the time it takes to copy data to the GPU memory for computing gradient estimates on workers.
On the other hand, we do not pin the test set to memory as using it is usually much less frequent than the training data. 
Second, we pin model weights in the parameter server main memory as gRPC currently does not allow communicating values that reside on the GPU memory.
Given that the model weights are communicated in each round, we never copy such weights to the GPU memory (except when testing the accuracy).
Yet, we store the model on workers on the GPU (or multiple GPUs whenever possible) so as to accelerate gradient computation.

\begin{figure*}[th!]
\centering 
\subfloat[Single server, multiple workers]{\includegraphics[width=0.32\linewidth,keepaspectratio]{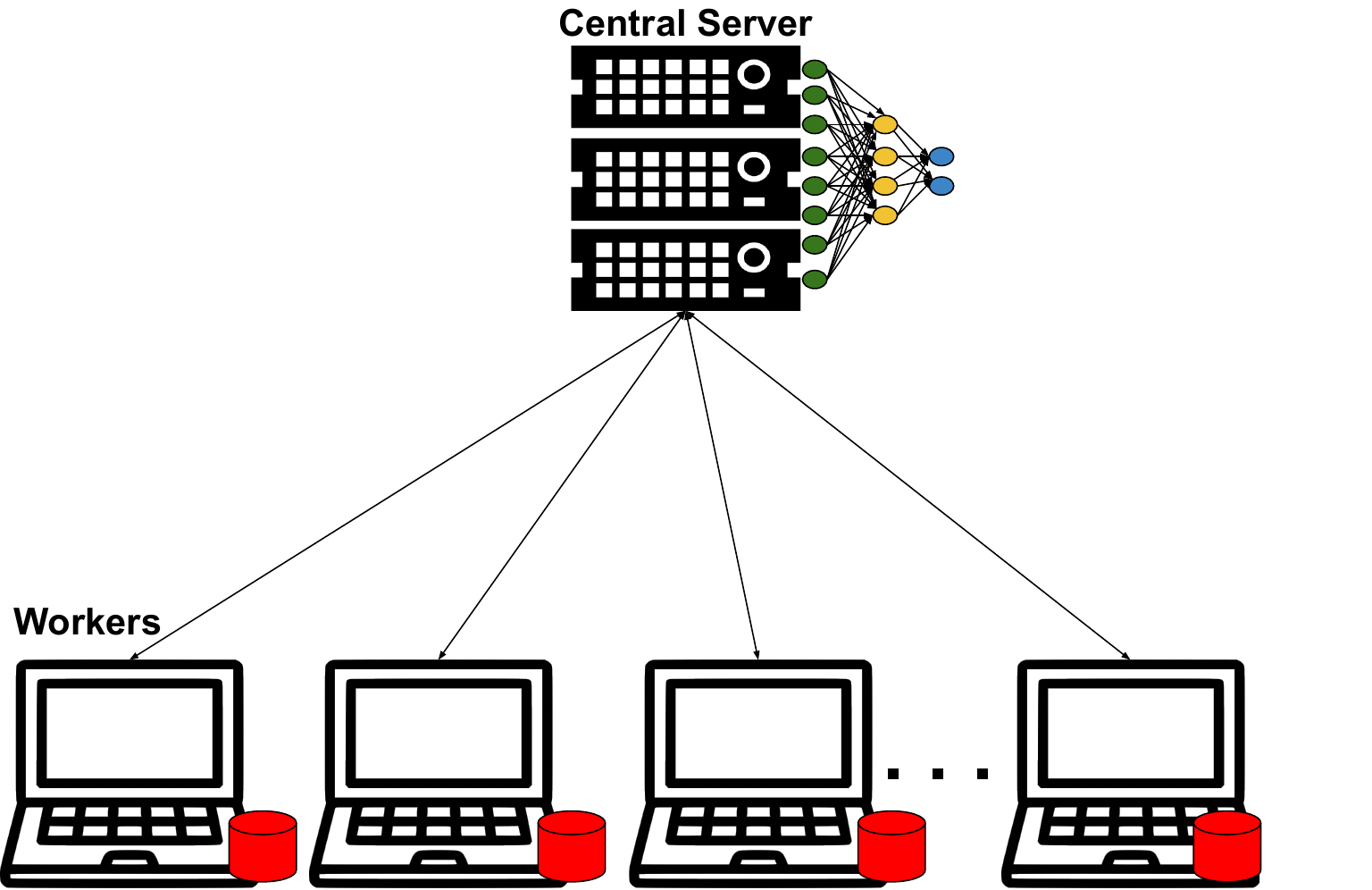}
\label{subfig:aggregathor_setup}}
\subfloat[Multiple server, multiple workers]{\includegraphics[width=0.32\linewidth,keepaspectratio]{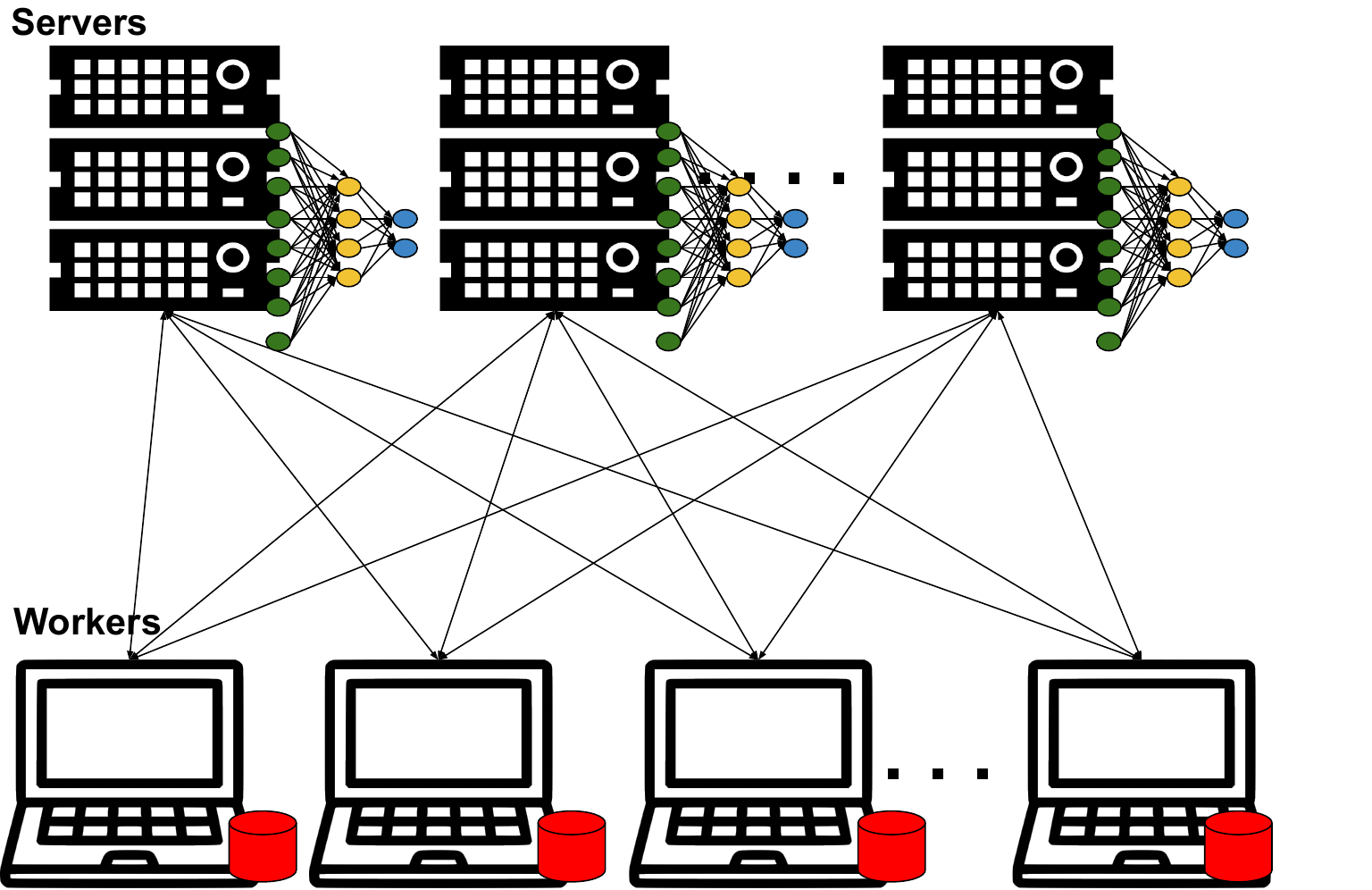}
\label{subfig:byzsgd_setup}}
\subfloat[Decentralized learning]{\includegraphics[width=0.32\linewidth,keepaspectratio]{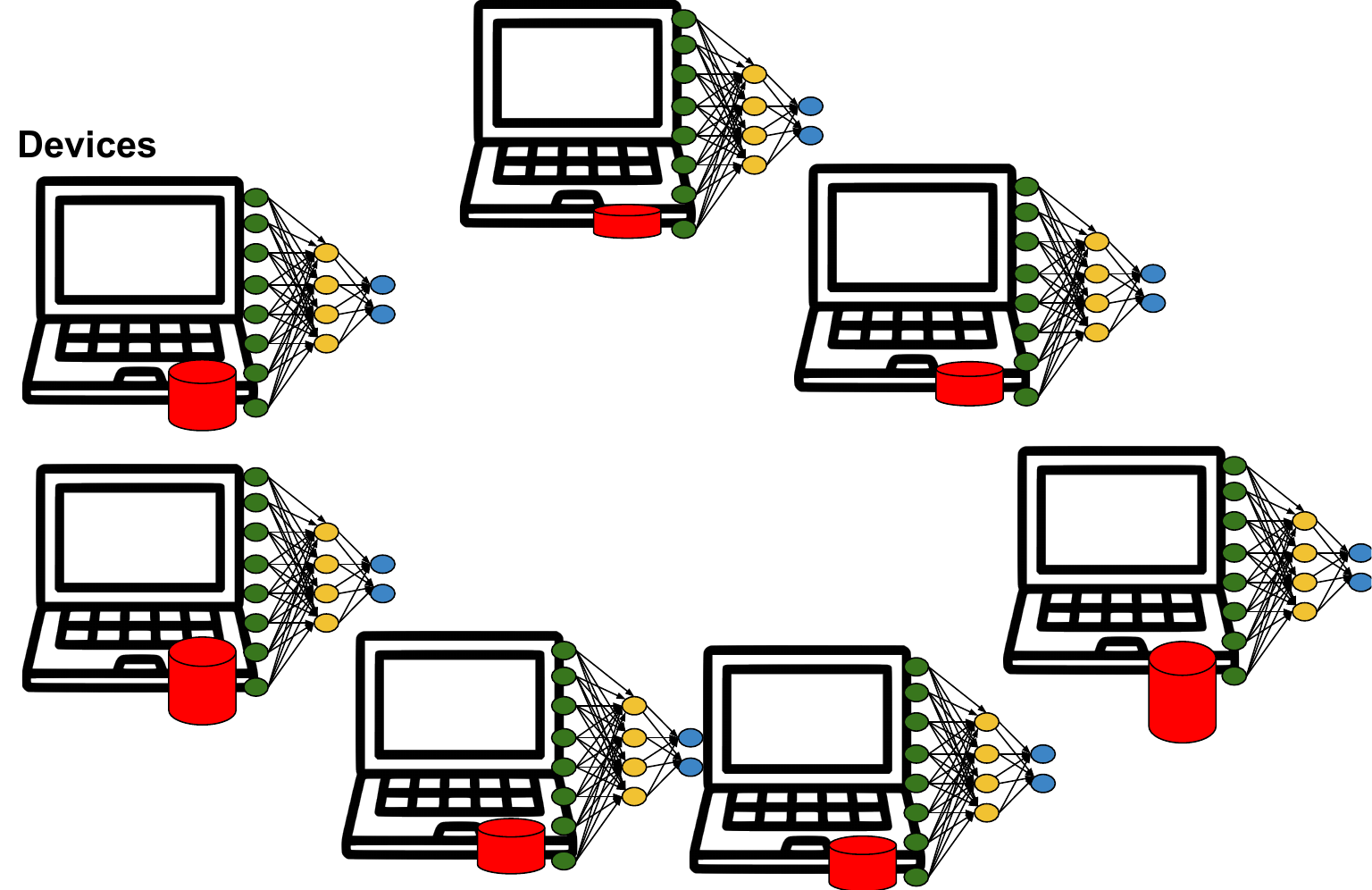}
\label{subfig:learn_setup}}
\caption{Examples of applications that can be built using \system{}.}
\label{fig:applications}
\end{figure*}

We carefully optimize memory usage by our GARs.
For example, aggregating gradients may require multiple iterations, calculating some distance-based scores for each of them in each iteration, e.g.,\ with \emph{Multi-Krum} or \emph{Bulyan}. 
We then cache the results of each of these iterations (in the CPU or GPU memory) and hence, remove redundant computations. Besides, we reduce the memory cost by allocating space only for one iteration along with the intermediate selected gradients.

\section{Applications}
\label{sec:applications}

In this section, we show how one can use \system{} to build Byzantine ML applications. 
We give three examples (depicted in Figure~\ref{fig:applications}) that span different architectures for Byzantine--resilient algorithms.

\subsection{Single Server, Multiple Workers (SSMW)}
Our first application represents the standard setup that was vastly studied in the last few years, e.g.,\ in~\cite{krum,alistarh2018byzantine,aggregathor}. Such setup uses the vanilla parameter server architecture~\cite{li2013parameter} (see Figure~\ref{subfig:aggregathor_setup}), yet with one crucial difference:
instead of aggregating the workers' updates by averaging them, the server uses a statistically--robust GAR (see Section~\ref{subsec:gars}) for aggregation.
Such setup usually assumes synchronous network, i.e.,\ there is an upper bound on the time it takes the workers to compute gradients and to reply to the server.

Listing~\ref{lst:aggregathor_app} shows how to build such setup, using \system{}, in a few lines of code.

\lstset{caption={Implementing \emph{SSMW} setup with \system{}.}}
\lstset{label={lst:aggregathor_app}}
\begin{lstlisting}
# parsing training arguments
ps = Server(...) #args omitted for brevity
for i in range(num_iter):
  gradients = ps.get_gradients(i, nw)
  aggr_grad = gar(gradients=gradients, f=fw)
  ps.update_model(aggr_grad)
  if i%comp_acc_freq == 0:
    acc = ps.compute_accuracy()
\end{lstlisting}

This depicts the code on the parameter server side, given the passiveness of workers in our design. 
First, the server object is initiated (line 2) with the appropriate parameters.
Then, the training loop (lines 3--6) runs as much as the user specifies.
In each training iteration, the server first asks the workers to compute gradient estimates using the current model state.
Note that, the second argument in \emph{get\_gradients()} method specifies the number of replies the server should wait for before continuing the iteration.
The server then applies some GAR on the received gradients and then updates the model using the aggregated gradient.
From now and then, the server computes the accuracy of the current model state (lines 7--8).

Notably, \aggregathor{} official code base~\cite{aggregathorcode} (which implements this setup) is in the order of hundreds of lines of code. 

\subsection{Multiple Server, Multiple Workers (MSMW)}
Our second application extends the first one by considering multiple servers and multiple workers.
This setup was considered  recently~\cite{el2020genuinely} to tolerate Byzantine servers as well as Byzantine workers.
Such a setup requires replicating
the server on multiple machines (see Figure~\ref{subfig:byzsgd_setup}).
It can also accommodate asynchronous networks in the sense that it does not assume any upper bound on the computation nor communication delays.

Listing~\ref{lst:byzsgd_app} shows how to build \emph{MSMW} setup, allowing multiple server replicas and asynchronous communication.

\lstset{caption={Implementing \emph{MSMW} setup with \system{}.}}
\lstset{label={lst:byzsgd_app}}
\begin{lstlisting}
# parsing training arguments
ps = Server(...) #args omitted for brevity
for i in range(num_iter):
  gradients = ps.get_gradients(i, nw-fw)
  aggr_grad = gar(gradients=gradients, f=fw)
  ps.update_model(aggr_grad)
  models = ps.get_models(nps-fps)
  aggr_models = gar(gradients=models,f=fps)
  ps.write_model(aggr_models)
  if i%comp_acc_freq == 0:
    acc = ps.compute_accuracy()
\end{lstlisting}

We focus here again on the server side.
Lines 1--6 are very similar to those in Listing~\ref{lst:aggregathor_app} (except for the number of expected gradients to collect in line 4).
Lines 7--9 show the additional communication step required among the server replicas to ensure convergence.
First, each server fetches the updated model from a few other servers (whose number is given as an argument to \emph{get\_models()} method).
Then, the collected models are aggregated, and the result is stored as the new model state.

\subsection{Decentralized Learning}
Our third application considers the Byzantine ML problem in decentralized settings as in e.g., \cite{yang2019byrdie,yang2019bridge,el2020collaborative}.
Decentralized learning is different from the previous two use cases as it does not use the parameter server architecture (see Figure~\ref{subfig:learn_setup}), i.e.,\ communication is done in a peer--to--peer fashion.
Such setup is mainly useful when data is sensitive and should be kept private.
Notably, it addresses settings where data is not identically distributed on the contributing machines.

Listing~\ref{lst:learn_app} shows how to build such an application, allowing for a decentralized setup and multiple communication steps per training iteration.

\lstset{caption={Implementing decentralized learning with \system{}.}}
\lstset{label={lst:learn_app}}
\begin{lstlisting}
# parsing training arguments
wrk = Worker(...) #args omitted for brevity
ps = Server(...) #args omitted for brevity
for i in range(num_iter):
  gradients = ps.get_gradients(i, n-f)
  aggr_grad = gar(gradients=gradients, f=f)
  if non_iid:
    aggr_grad = contract(...)
  ps.update_model(aggr_grad)
  models = ps.get_models(n-f)
  aggr_models = gar(gradients=models,f=f)
  ps.write_model(aggr_models)
  if i%comp_acc_freq == 0:
    acc = ps.compute_accuracy()

def contract(...):
  for _ in range(steps):
    ps.latest_aggr_grad = aggr_grad
    aggr_grads = ps.get_aggr_grads(n-f)
    aggr_grad = gar(gradients=aggr_grads, f=f)
  return aggr_grad

\end{lstlisting}

We highlight here the two main differences with the previous examples.
First, each node creates both a Server and a Worker objects (lines 2--3).
Second, there is a multi--round step (lines 16--21) in each iteration to \emph{contract} models on correct machines, especially when the data is not identically distributed on them.
The goal of this step is to force the model states on all machines to get closer to each other.

\section{Performance Evaluation}
\label{sec:eval}
We first describe the settings we use and the baselines we consider.
We then show micro-benchmarks for the GARs we implemented, followed by large--scale experiments, evaluating \system{} using the applications given in the previous section.

\subsection{Setting}
\paragraph{Testbed}
Our experimental platform is Grid5000~\cite{g5k}. For the experiments deployed with CPUs, we employ nodes from the same cluster, each having 2 CPUs (Intel Xeon E5-2630 v4) with 10 cores, 256~GiB RAM and 2$\times$10~Gbps Ethernet. For the GPU-based experiments, we employ nodes from two clusters (due to the limited number of nodes in one cluster); nodes in different clusters have different specifications. Each node has 2 identical GPUs.

\paragraph{Metrics} We use the following standard metrics:

\noindent{\it Accuracy.} This measures the top-1 cross-accuracy: the fraction of correct predictions among all the predictions using the \emph{test} set; 
this shows the quality of the learned model over time. 

\noindent{\it Throughput.} This quantifies the number of updates that the system processes per second. 
For deployments that employ multiple parameter servers, we report the highest throughput, which corresponds to the fastest correct machine. 

 

\paragraph{Application} We consider the 3 applications discussed in Section \ref{sec:applications}.
Yet, in some experiments, we focus more on the setup with multiple servers and multiple workers (\byzsgd{}) as it gives the flexibility to test with different number of Byzantine servers and workers.
We consider two variants of \byzsgd{}: the first uses Bulyan~\cite{bulyanPaper} 
to aggregate gradients, and hence achieves Byzantine resilience in high dimensions while assuming network asynchrony. 
The second one uses \mkrum{}~\cite{aggregathor} to aggregate gradients while assuming network synchrony. 
Unless otherwise stated, we use our TensorFlow version with the first variant and PyTorch for the second one. 

We consider 
image classification 
due to its wide adoption as a benchmark for distributed ML systems~\cite{chilimbi2014project,abadi2016tensorflow}. 
We use MNIST~\cite{mnist} and CIFAR-10~\cite{cifar} datasets. MNIST is a dataset of handwritten digits with 70,000 $28\times28$ images in 10 classes. 
CIFAR-10 consists of 60,000 $32\times32$ colour images in 10 classes. Table~\ref{table:models} presents the models we use for evaluation.

\begin{table}
\caption{Models used to evaluate \system{}.}
\label{table:models}
\vspace{-2mm}
\centering
\begin{tabular}{|c|c|c|}
\hline
\textbf{Model}         & \# \textbf{parameters} & \textbf{Size (MB)} \\ \hline
MNIST\_CNN    & 79510         & 0.3       \\ \hline
CifarNet      & 1756426       & 6.7       \\ \hline
Inception  & 5602874       & 21.4      \\ \hline
ResNet-50  & 23539850      & 89.8      \\ \hline
ResNet-200 & 62697610      & 239.2     \\ \hline
VGG        & 128807306     & 491.4     \\ \hline

\end{tabular}
\vspace{-4mm}
\end{table}


\paragraph{Setup} For TensorFlow experiments, we employ 18 workers, out of them 3 could be faulty ($n_w=18,f_w=3$) and 6 servers, 1 could be faulty ($n_{ps}=6,f_{ps}=1$). Note that 
in \learn{} experiments, we do not use any servers (i.e.,\ workers communicate in a peer--to--peer fashion). We employ a batch size of 32 at each worker, leading to an effective batch size of 480 in the normal case. 
Recent studies show that going further than this number does not help achieve faster convergence~\cite{shallue2018measuring} in addition to reducing throughput (due to the computation overhead). 
For PyTorch experiment, we use 10 workers, with also 3 Byzantine, and 3 servers, with only 1 Byzantine. We use a batch size of 100 at each worker. 
We repeated all the experiments multiple times and found that the error-bars are always very small compared to the presented values and hence, we omit them for better readability.

\subsection{Baselines}
To the best of our knowledge, \system{} is the first library to 
accommodate 
both Byzantine servers and workers. 
We chose the following baselines to 
compare \system{} to.
\paragraph{Vanilla baseline}
This is the vanilla deployment of TensorFlow or PyTorch.
Such deployment fails to tolerate any Byzantine behavior whatsoever.
Comparing \system{} against this baseline quantifies the overhead of 
Byzantine resilience. 

\paragraph{\aggregathor~\cite{aggregathor}} This is the only
existing scalable ML system that achieves Byzantine resilience, yet only for Byzantine workers.
It is built on TensorFlow and supports training only on CPUs. 
\aggregathor{} uses one central, trusted server while tolerating Byzantine workers (i.e.,\ uses \textit{SSMW} setup), and it considers synchronous networks.
For a fair comparison with our \system{}--based systems, we use the same GAR for both deployments.
Thus, comparing with this baseline quantifies the overhead of using our object--oriented design and the communication layer we provide.

\paragraph{Crash--tolerant protocol} We implement a strawman approach to tolerate crash failures, assuming synchronous communication, using \system{} components. As worker crashes do not affect the learning convergence eventually, we only tolerate server crashes, by replicating the server. Server replicas 
get the updates from all workers and \emph{average} them in each iteration, but workers contact only one of these replicas, i.e.,\ the \emph{primary}, to get the updated model. In the case of \emph{primary} crash (signaled by a timeout), workers contact the \emph{next} server, marking it as the new \emph{primary}. The new primary sends its view of the model to all workers so as to inform them about the change. The model sent by the new primary could be outdated compared to the model of the crashed primary (due to missing some updates). This is still fine and learning will converge eventually~\cite{xie2018faster}, given that $n_{ps}\ge f_{ps}+1$, where $n_{ps}$ is the total number of replicas, and $f_{ps}$ is the maximum number of crashing nodes, i.e.,\ servers. Thus, this deployment guarantees eventual convergence without any guarantees on throughput nor convergence rate.
Some ML systems already use Paxos~\cite{lamport2001paxos} for crash fault tolerance~\cite{chilimbi2014project,li2013parameter}. However, our strawman algorithm, we believe, gives strictly weaker guarantees (in terms of consistency of model state among replicas), and hence has a better throughput than Paxos.\footnote{Unfortunately, there is no open-source code for direct comparison.} 

\begin{figure}[t]
\centering
\vspace{-2mm}
\subfloat[Number of Inputs]{\includegraphics[width=0.49\linewidth,keepaspectratio]{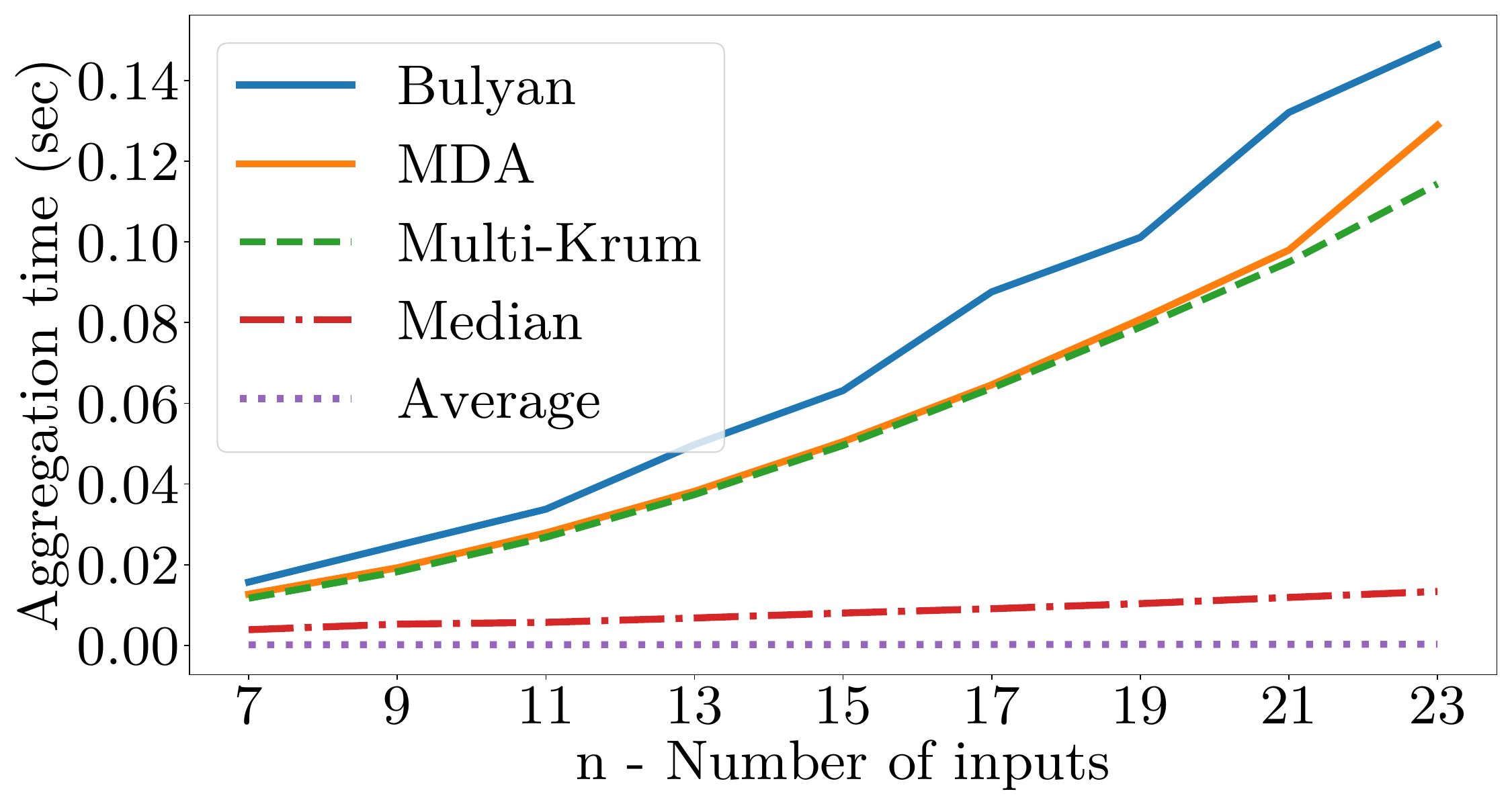}
\label{subfig:microbench-n}}
\subfloat[Input dimension]{\includegraphics[width=0.49\linewidth,keepaspectratio]{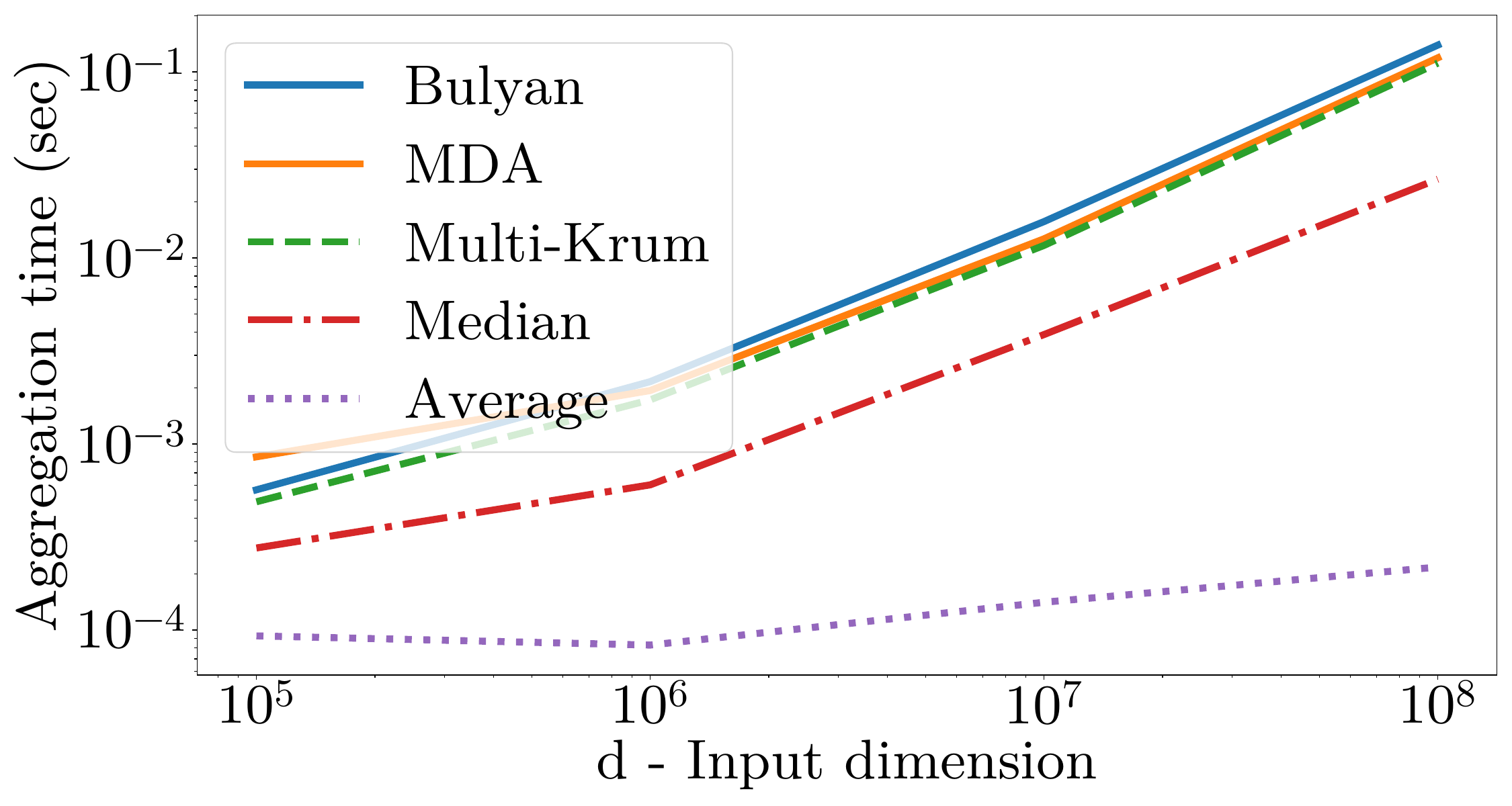}
\label{subfig:microbench-d}} 
\caption{Micro-benchmark of different GARs deployed on a GPU. $n$ denotes the number of inputs to the GAR and $d$ denotes the length of one input/gradient.}
\label{fig:microbench-tf}
\vspace{-4mm}
\end{figure}

\begin{figure*}[!ht]
\centering 
\vspace{-4mm}
\subfloat[Convergence with CifarNet]{\includegraphics[width=0.49\linewidth,keepaspectratio]{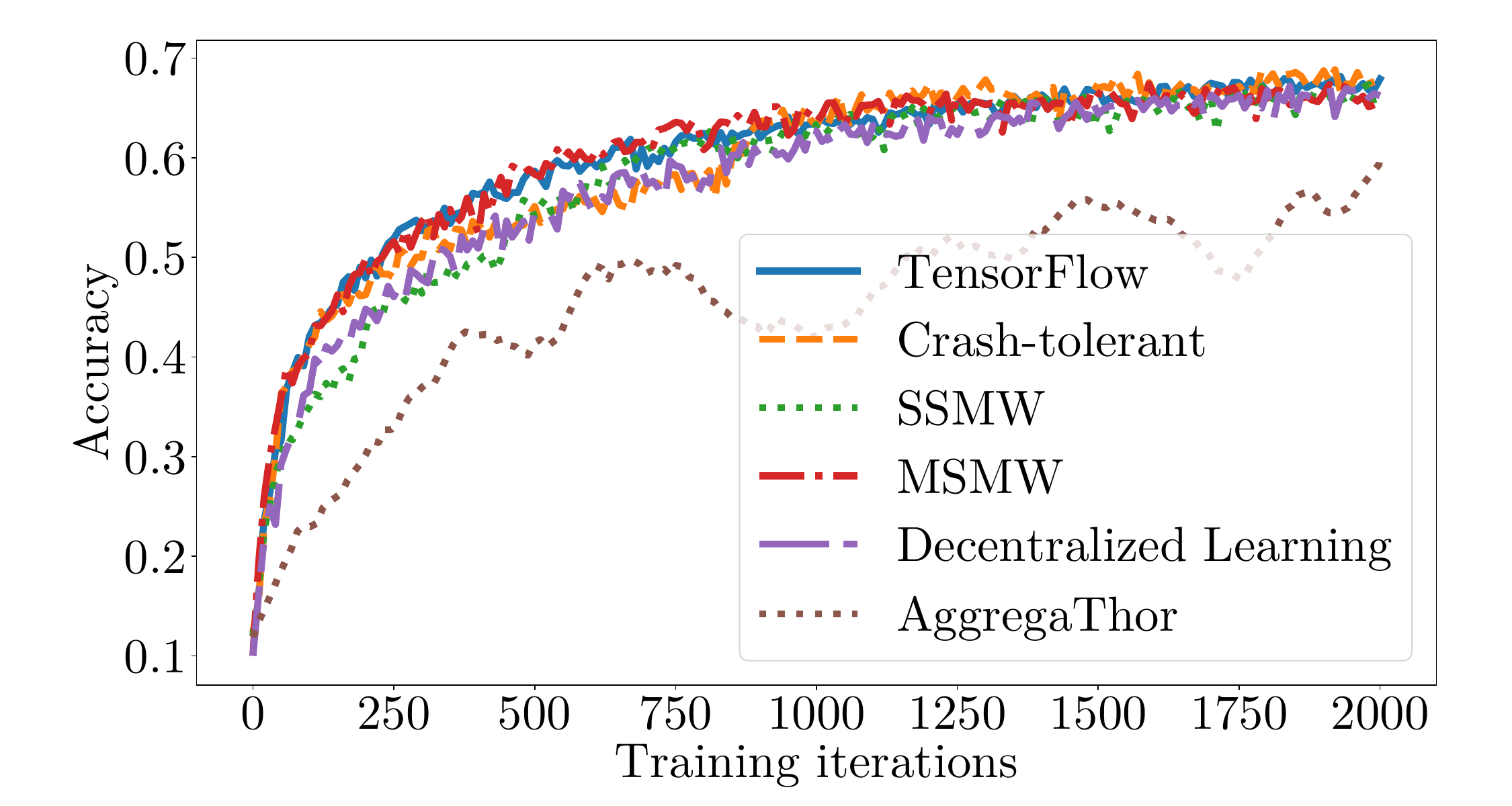}
\label{subfig:conv_tf}}
\subfloat[Convergence with ResNet-50 (1 epoch = 200 iterations)]{\includegraphics[width=0.49\linewidth,keepaspectratio]{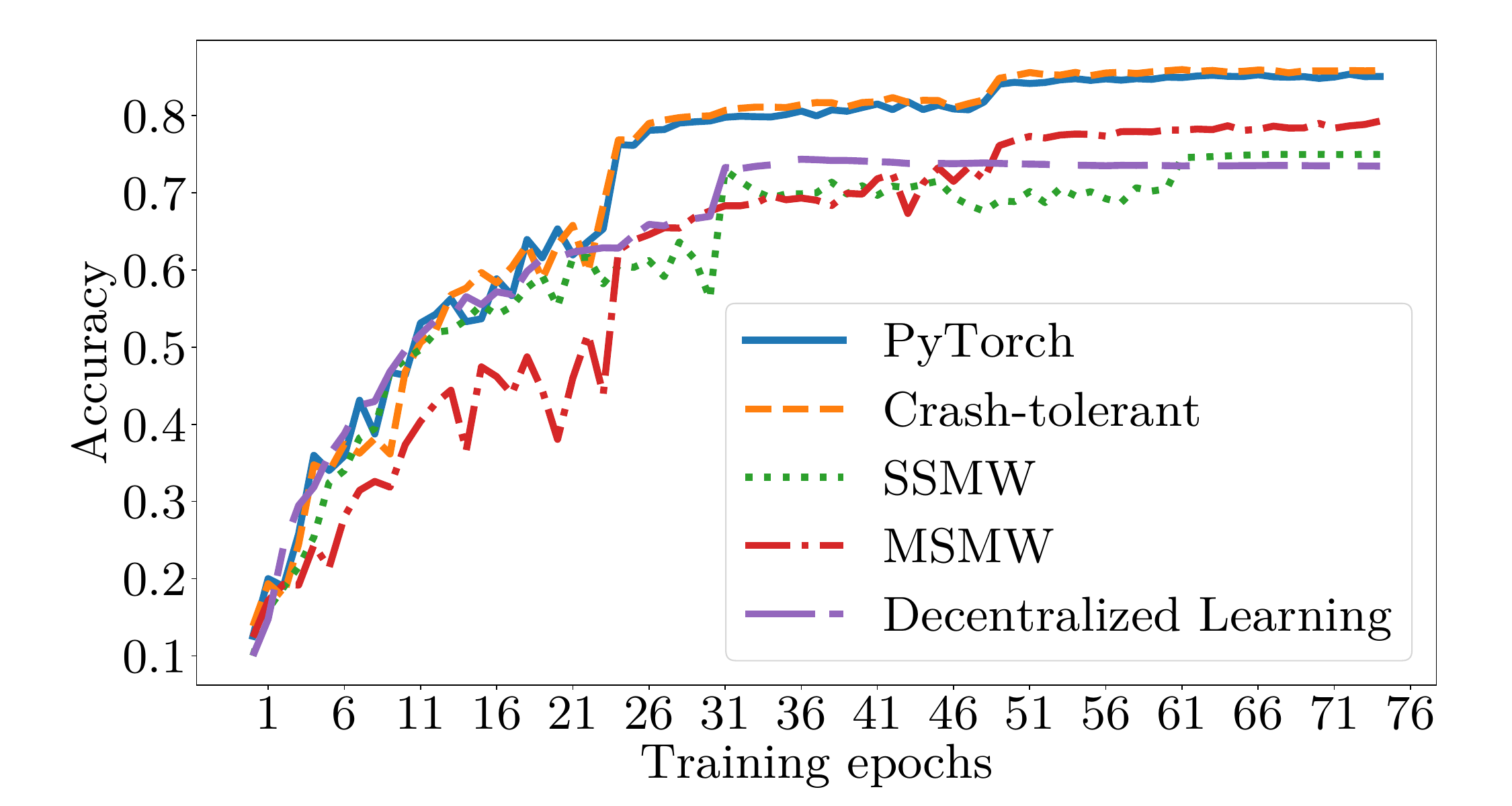}
\label{subfig:conv_pt}}
\vspace{-2mm}
\caption{Convergence of \system{} applications with respect to other baselines using two models.}
\vspace{-4mm}
\label{fig:conv-tf}
\end{figure*}

\subsection{GARs Micro--benchmarks}
The Byzantine--resilient GARs are the basic enabling tools for Byzantine resilience.
We provide a micro--benchmark for their GPU--based implementation performance with respect to the number of inputs/gradients (i.e.,\ $n$ the number of workers or servers) and the gradient dimension (i.e.,\ $d$).


Vanilla frameworks, e.g.,\ TensorFlow, \emph{average} gradients at the parameter server. 
So, we also include the evaluation of \emph{Average}, which has been implemented as a part of the \system{} library. 
\emph{Average} is our baseline in this experiment.

For a fair comparison, we set $f$, the number of declared Byzantine inputs, to $\left \lfloor{\frac{n-3}{4}}\right \rfloor$ for all Byzantine-resilient GARs and hence, the smallest possible $n$ is 7.
We set $d=10^7$ in Figure~\ref{subfig:microbench-n} and $n=17$ in Figure~\ref{subfig:microbench-d}.
The metric for this micro--benchmark is the \emph{aggregation time}:
it includes the aggregation of $n$ input vectors (all resident in GPU memory)
and the transfer of the resulting vector back to main memory.
Each point is the average of $21$ runs, for which we observed a standard deviation two orders of magnitude below the observed average.
We ran this micro--benchmark on an \emph{Intel Core i7-8700K} CPU and two \emph{Nvidia GeForce 1080 Ti} GPUs.

Theoretically, the asymptotic complexities of \brute{}, \mkrum{}, \bulyan{}, and \emph{Average} are respectively $\mathcal{O}\!\left( \binom{n}{f} + n^2 d \right)$ \cite{bulyanPaper}, $\mathcal{O}\!\left( n^2 d \right)$~\cite{krum}, $\mathcal{O}\!\left( n^2 d \right)$~\cite{bulyanPaper} and $\mathcal{O}\!\left( n d \right)$.
Our implementation of \medianoid{} has a best case complexity of $\mathcal{O}\!\left( n d \right)$ and worst case of $\mathcal{O}\!\left( n^2 d \right)$.
In practice, for a fixed $d$ (Figure~\ref{subfig:microbench-n}), we observe these asymptotic behaviors for \mkrum{} and \bulyan{}: quadratic in $n$. 
\medianoid{} shows good scalability with $n$, maintaining a consistent performance that is very close to \emph{Average}. 
Although the asymptotic complexity of \brute{} is exponential, our implementation achieves only a quadratic growth with $n$.
The values of $n$ and $f$ used in these experiments are merely too low to expose such a behavior, i.e.,\ the exponential growth with $n$.
\emph{Average} aggregation time remains roughly constant for a fixed $d$ and $n < 15$, with an aggregation time of $\sim{}\!8\text{ ms}$, and then grows linearly. 
For a fixed $n$ (Figure~\ref{subfig:microbench-d}), we observe a linear time increases with respect to $d$ for every one of the studied GARs.


\subsection{Convergence Comparison}
\label{subsec:conv}

Figure~\ref{fig:conv-tf} shows the results of two experiments: the first one (Figure~\ref{subfig:conv_tf}) trains CifarNet on TensorFlow--based systems (including \aggregathor{}) 
using CPUs, where the second one (Figure~\ref{subfig:conv_pt}) trains ResNet-50 on PyTorch--based systems using GPUs. Both experiments use CIFAR-10 as a dataset. 
The first experiment puts \system{} in a head--to--head comparison with the state--of--the--art Byzantine ML system, i.e., \aggregathor. 
The second experiment is an instance of a deployment of \system{} while training a bigger model using GPUs. 


Figure~\ref{subfig:conv_tf} shows that all the systems achieve almost the same final accuracy (except \aggregathor{}). 
Some of the Byzantine--resilient deployments converge a bit slower than those using \emph{averaging} during training, yet reaching the same accuracy 
eventually, i.e.,\ after doing enough number of iterations.
Interestingly, we can notice that Byzantine--resilient applications do not add much overhead compared to the crash--tolerant one, in terms of the number of iterations till convergence (less than $1\%$).
Surprisingly, \system{} applications achieve better final accuracy than \aggregathor{}.
We speculate that the reason is the fact that \aggregathor{} relies on an old version of TensorFlow compared to \system{} (1.10 vs. 2.3).
Another related reason is the fact that the latest version of TensorFlow is also integrated with the highly--optimized \emph{Keras} library, which we also use.
Figure~\ref{subfig:conv_pt} shows that Byzantine--resilient applications fail to reach the same final accuracy as vanilla PyTorch, with up to $10\%$ final accuracy loss. 
This accuracy loss, 
although not clear in Figure~\ref{subfig:conv_tf}, makes sense as a direct byproduct of using Byzantine--resilient GARs to aggregate workers' gradients and also due to having diverging servers (in some cases).
We notice that combining network asynchrony with decentralization leads to the biggest accuracy loss.  
Asynchrony essentially leads to the aggregation of outdated models and gradients, slowing down convergence and reducing the final accuracy.
Interestingly, the crash--tolerant deployment does not experience such a loss compared to the vanilla case.

\begin{figure}[!t]
\centering 
\vspace{-4mm}
\subfloat[Random vectors]{\includegraphics[width=0.49\linewidth,keepaspectratio]{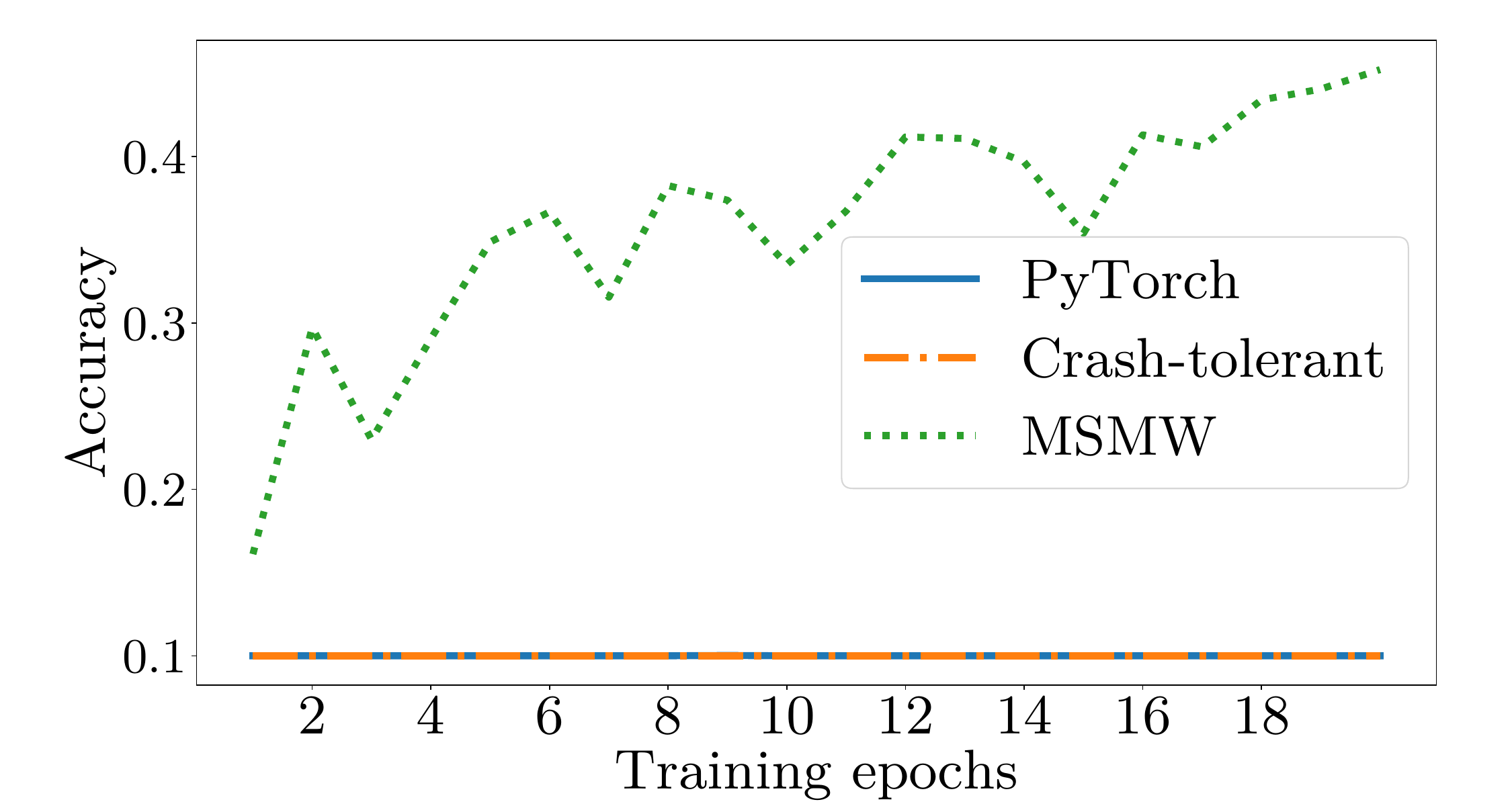}
\label{subfig:byz_rand}}
\subfloat[Reversed vectors]{\includegraphics[width=0.49\linewidth,keepaspectratio]{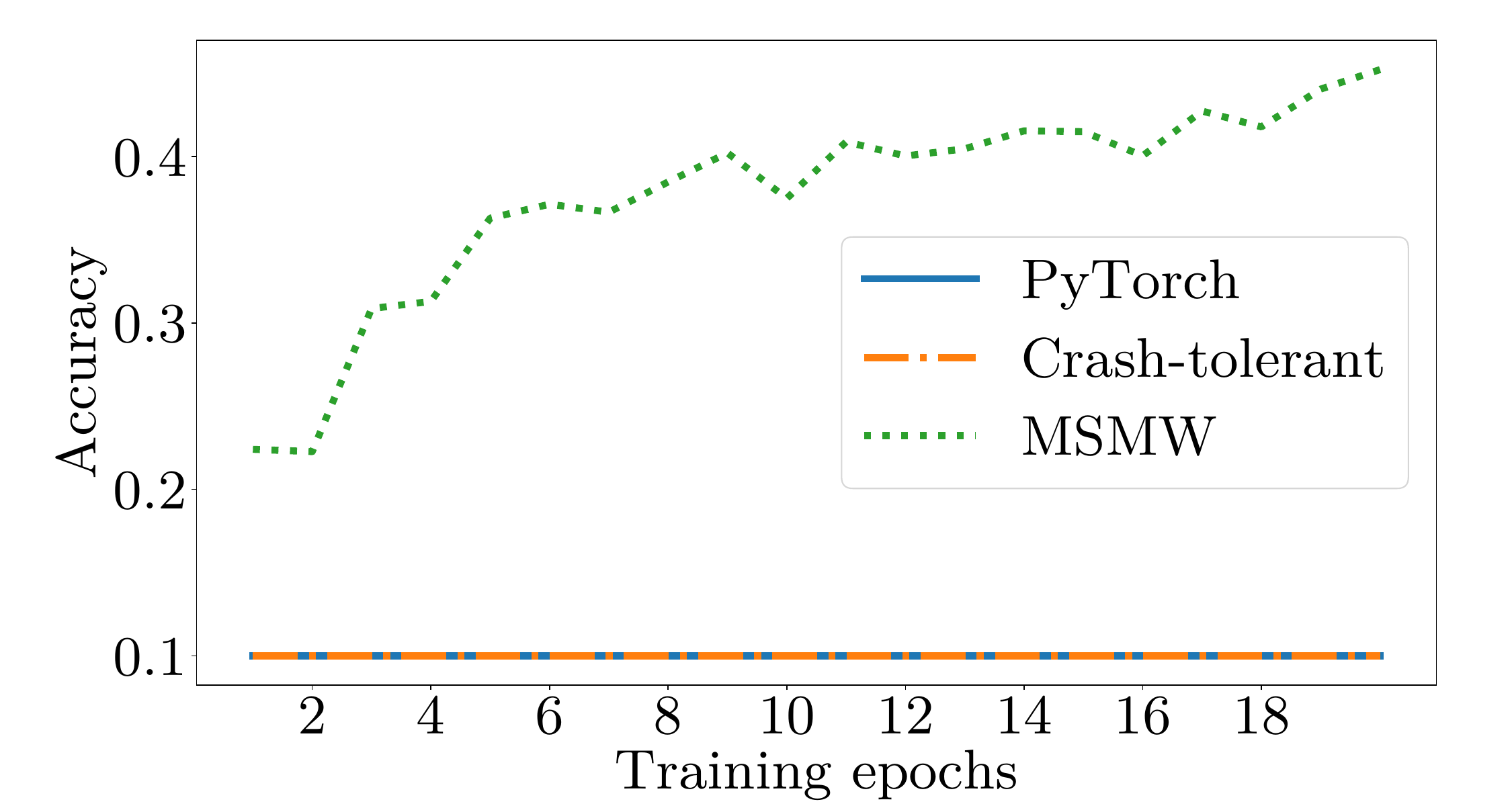}
\label{subfig:byz_min}}
\caption{\system{} tolerance to two Byzantine attacks.}
\vspace{-6mm}
\label{fig:byz}
\end{figure}

\subsection{Byzantine Behavior}
\label{subsec:byzexp}
As a sanity check to our implementation, we conduct experiments with real Byzantine behavior, where we apply attacks on the vanilla baseline (PyTorch in this experiment), a \system{}--based application (\byzsgd{} in this experiment), and the crash--tolerant protocol. Figure \ref{fig:byz} shows two kinds of attacks 
on both servers and workers sides. In the first attack (Figure \ref{subfig:byz_rand}), the Byzantine node, be it a worker or a server, replaces its data, be it a gradient or a model, with random values, where in the second attack (Figure~\ref{subfig:byz_min}), such vectors are reversed and amplified (multiplied by -100). We train CifarNet with 11 workers and 3 servers (in the case of fault--tolerant algorithms) with 1 Byzantine node from each party. We do the training only for 20 epochs rather than till convergence. On the one hand, both the vanilla deployment and the crash--tolerant deployment fail to learn under both attacks. On the other hand, \byzsgd{} manages to train the model safely and converges to a normal, high accuracy. 

\subsection{Throughput}
We show here the computation and the communication costs of Byzantine resilience. 
First, we quantify the overhead of employing \system{}--based applications compared to the other baselines by measuring the throughput while training several models. 
Then, we analyze the scalability of such applications with a different number of workers, Byzantine workers, and Byzantine servers. 
In this section and without loss of generality, we use ResNet-50 as our model (unless otherwise stated). 
We do not employ any attack nor Byzantine behavior in these experiments as we want to quantify the overhead of Byzantine resilience in a normal, optimistic environment. Thus, we denote here the number of \emph{declared} Byzantine nodes 
with the number of Byzantine nodes (i.e.,\ $f_w$ and $f_{ps}$).

\begin{figure}[!t]
\centering
\vspace{-2mm}
\subfloat[CPU]{\includegraphics[width=0.99\linewidth,keepaspectratio]{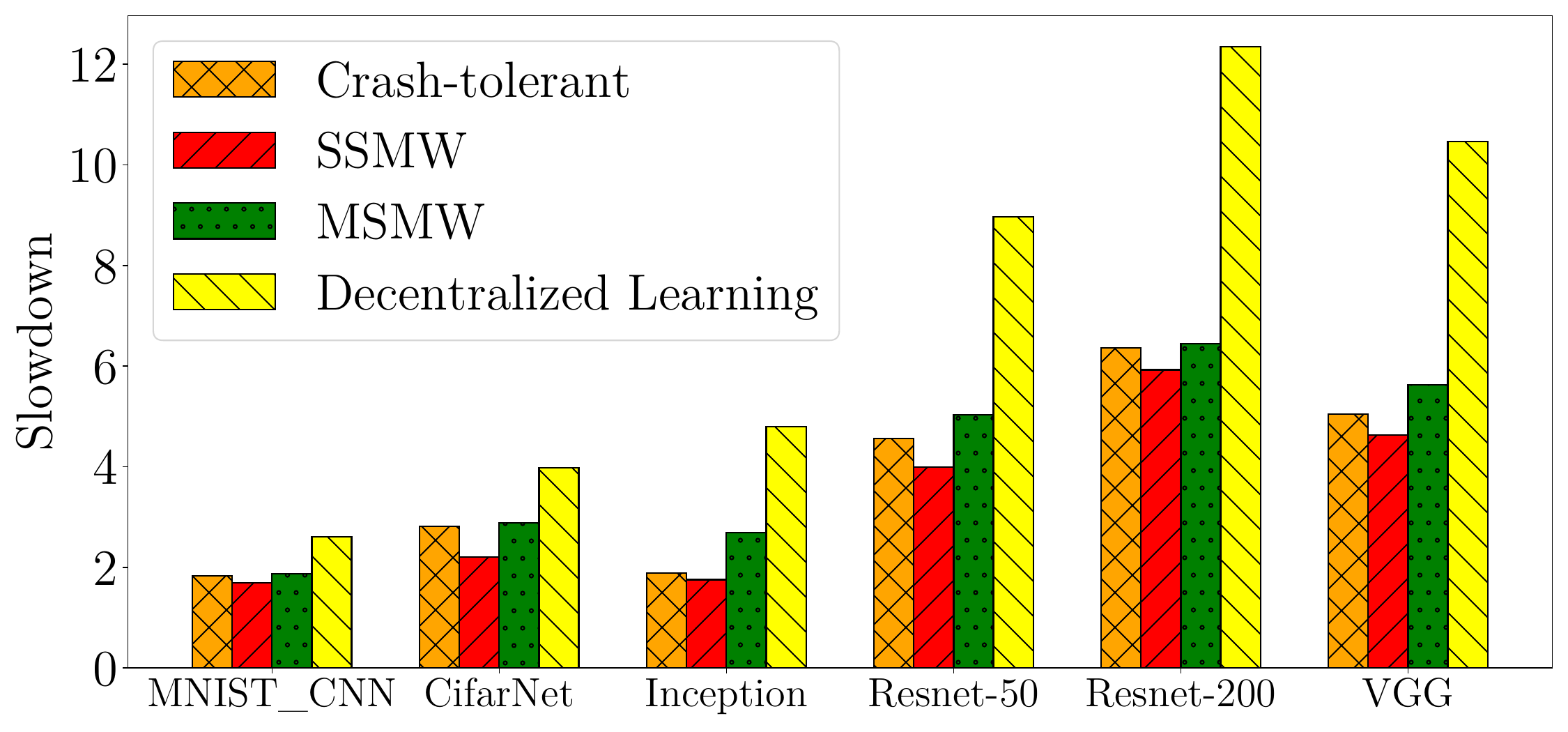}
\label{subfig:throughput-cpu}}
\\\vspace{-4mm}
\subfloat[GPU]{\includegraphics[width=0.99\linewidth,keepaspectratio]{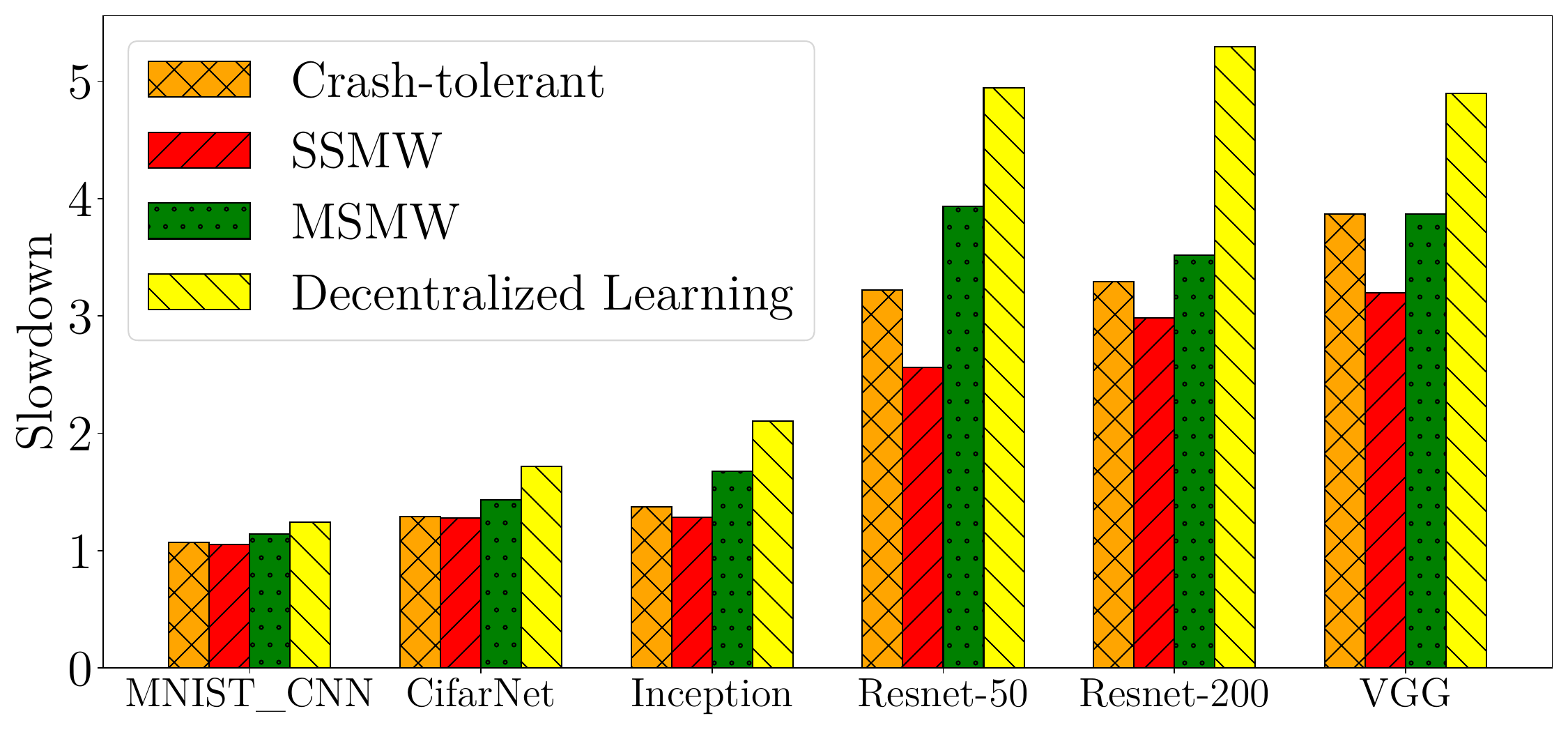}
\label{subfig:throughput-gpu}} 
\vspace{-2mm}
\caption{Slowdown of fault--tolerant systems normalized to the vanilla baseline (i.e.,\ TensorFlow) throughput.}
\label{fig:overhead-tf}
\vspace{-6mm}
\end{figure}

\paragraph{Model dimension}
Figure~\ref{fig:overhead-tf} depicts the cost of Byzantine resilience in terms of throughput.
The throughput of the fault--tolerant systems is normalized to the vanilla baseline throughput in each case. Thus, the y-axis represents the slowdown that each of the fault--tolerant systems induces compared to the vanilla baseline.  
The overhead of crash tolerance ranges from $83\%$ to $537\%$ ($7\%$--$286\%$), that of \textit{SSMW} ranges from $69\%$ to $492\%$ ($5\%$--$219\%$), that of \byzsgd{} ranges from $88\%$ to $544\%$ ($14\%$--$292\%$), and that of \learn{} ranges from $161\%$ to $1135\%$ ($24\%$--$429\%$) compared to the vanilla deployments on CPUs (and GPUs). 
More interestingly, compared to the crash--tolerant deployment, \byzsgd{} overhead ranges from $1\%$ to $42\%$ ($0.1\%$--$22\%$) and \learn{} overhead ranges from $41\%$ to $154\%$ ($16\%$--$61\%$).
It is evident that CPU--based deployments show higher slowdowns than that of the GPU--based ones. 
We root that to two reasons: (1)~we test with more machines in the first case, inducing higher communication overhead, and (2)~GARs overhead is bigger with CPUs than with GPUs.

\begin{figure}[t]
\centering
\includegraphics[width=0.45\textwidth]{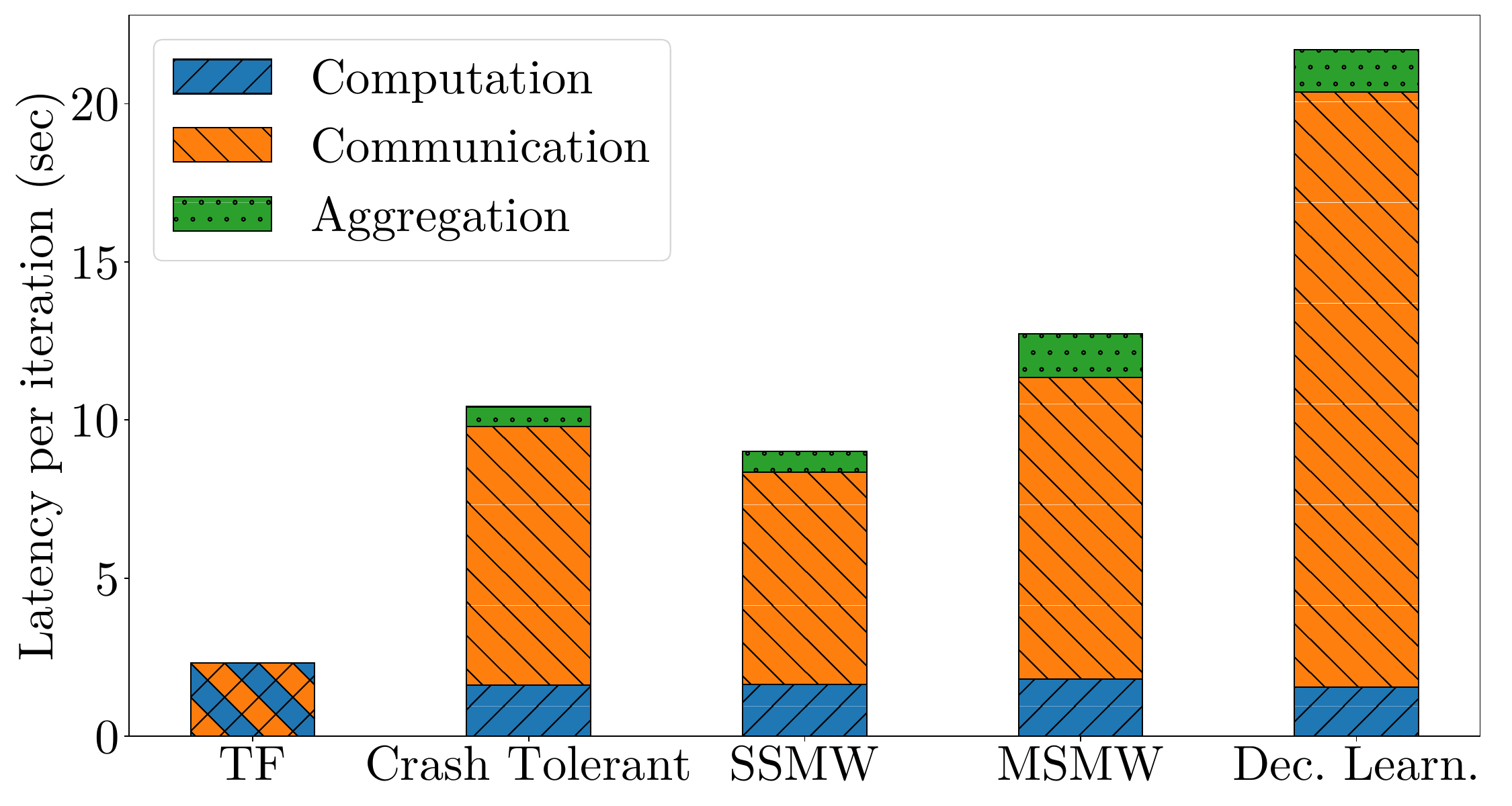}
\caption{Overhead breakdown in a CPU-based experiment.}
\label{fig:overbreak}
\vspace{-4mm}
\end{figure}

We extract several observations from Figure~\ref{fig:overhead-tf}. 
First, the cost of Byzantine resilience, compared to vanilla baselines, seems big (reaches $\sim$ 12x in the worst case) however, such a cost is reasonable compared to weaker alternatives (e.g.,\ crash tolerance).
Interestingly, the cost of workers' Byzantine resilience (using \textit{SSMW}) is always less than that of crash tolerance (more clear with big models).
Second, ML training, especially on GPUs, is network--bound: 
applications that require more communication have bigger slowdowns.
Essentially, communication constitutes more than $75\%$ of the overhead where,
robust aggregation 
account for less than $11\%$ 
(see Figure~\ref{fig:overbreak}). 
Third, increasing the model dimension increases the overhead of Byzantine resilience yet only until a certain point; after that point, the overhead remains roughly constant even with bigger models. 
The reason for that lies in the factors driving such an overhead. 
With small models, the bigger the model, the higher the cost of robust aggregation, 
the higher the overhead. 
Yet, with bigger models, the communication overhead prevails, which is in $\mathcal{O}(d)$ for all deployments. 

\paragraph{Overhead breakdown}
We pick one instance and take a closer look at all the deployments to understand the factors affecting their performance. Concretely, we run the same experiment while training ResNet-50, breaking the average latency per iteration for each deployment.
Figure~\ref{fig:overbreak} depicts the breakdown of the systems overhead when deployed on the CPU--based cluster. 
It is hard to decompose communication and the computation time for TensorFlow. 
Thus, the \emph{blue-and-orange} bar denotes the time spent in both of them 
combined. 

We can observe in the figure that the computation time is roughly the same for all applications ($\sim$ 1.6s).
Yet, the communication cost dominates the overhead (ranges from $75\%$ to $86\%$).
This makes (1) crash tolerance costly more than Byzantine workers' tolerance ($22\%$ extra communication), and (2) Byzantine servers' tolerance costly more than only workers' tolerance ($42\%$ more communication).
Furthermore,
we note that the aggregation time in \learn{} is two times bigger than that of \textit{SSMW}, due to the extra model aggregation step done by the former application.

\begin{figure}[!th]
\centering
\vspace{-6mm}
\subfloat[CPU TF-version]{\includegraphics[width=0.49\linewidth,keepaspectratio]{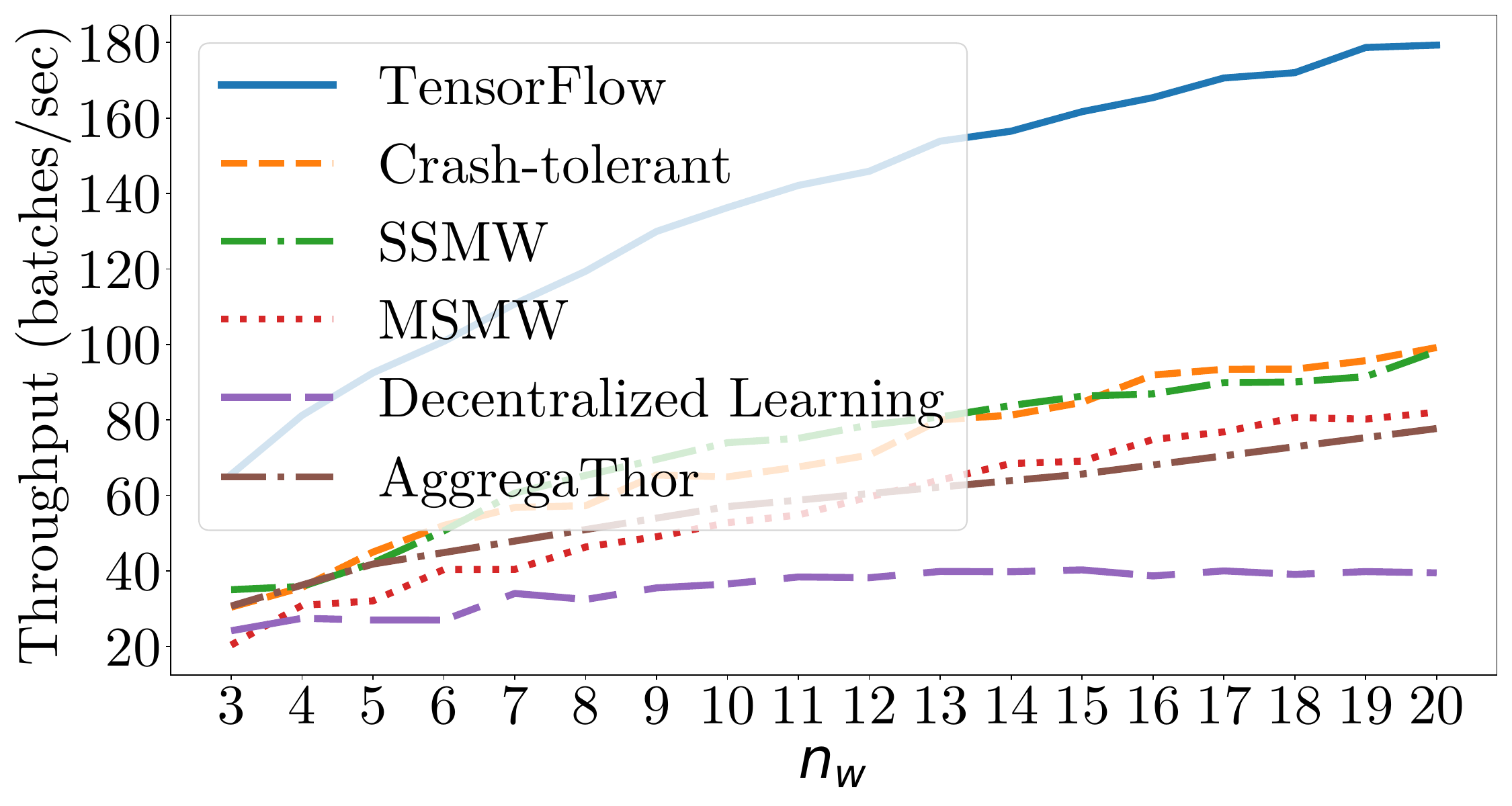}
\label{subfig:throughput-n-cpu}}
\subfloat[GPU PT-version]{\includegraphics[width=0.49\linewidth,keepaspectratio]{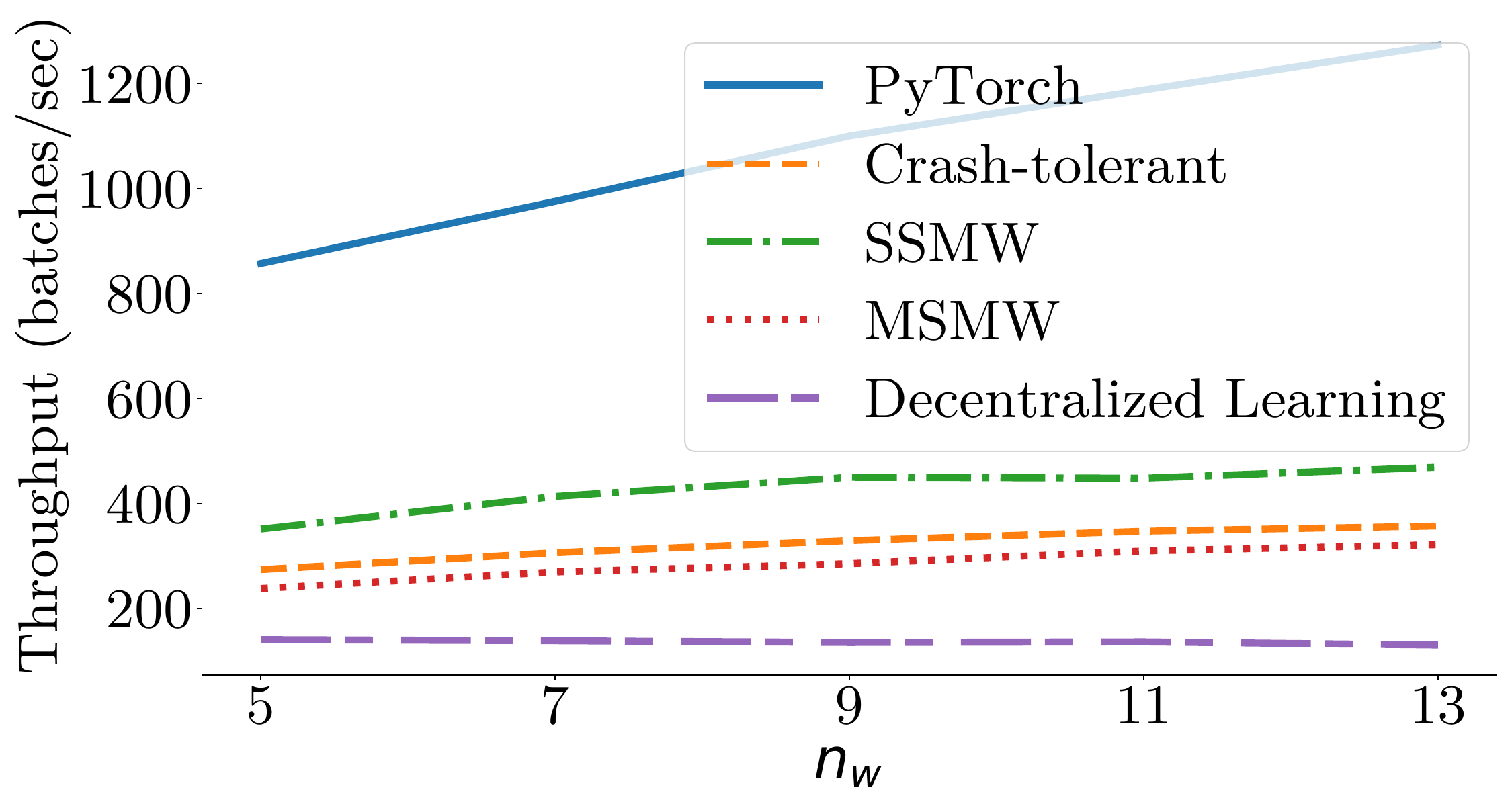}
\label{subfig:throughput-n-gpu}} 
\caption{Throughput comparison with increasing $n_w$.}
\label{fig:throughput-n}
\vspace{-4mm}
\end{figure}


\paragraph{Number of workers}
Increasing the number of workers ($n_w$), and hence increasing the \emph{effective} batch size, is crucial for scaling distributed ML applications.  
Figure~\ref{fig:throughput-n} depicts the scalability of \system{}--based applications 
while training CifarNet on CPUs (Figure~\ref{subfig:throughput-n-cpu}) and ResNet-50 on GPUs (Figure~\ref{subfig:throughput-n-gpu}). 
In this figure, throughput is measured in \emph{batches/sec} rather than \emph{updates/sec} since employing more workers allows for increasing the number of batches processed per iteration.

Figure~\ref{subfig:throughput-n-cpu} shows that \textit{SSMW} outperforms \aggregathor{}. This happens arguably due to the optimizations we include in \system{} in addition to using a newer version of TensorFlow.
We draw three main observations from Figure~\ref{fig:throughput-n}.
First, all systems 
scale with employing more workers (except the \learn{} application),
with around one order of magnitude higher throughput with GPUs compared to CPUs.
Second, the throughput gap between the vanilla deployments and the fault-tolerant deployments increases with increasing $n_w$, keeping the slowdown introduced by \textit{SSMW}, \byzsgd{}, and crash--tolerant almost constant.
Third, the scalability of \byzsgd{} is almost as good as that of the crash--tolerant deployment, and the difference in throughput with increasing $n_w$ is almost constant. This shows that \emph{complete} Byzantine resilience does not harm scalability compared to crash resilience.


\begin{figure}[th]
\centering
\vspace{-4mm}
\subfloat[Number of Inputs]{\includegraphics[width=0.49\linewidth,keepaspectratio]{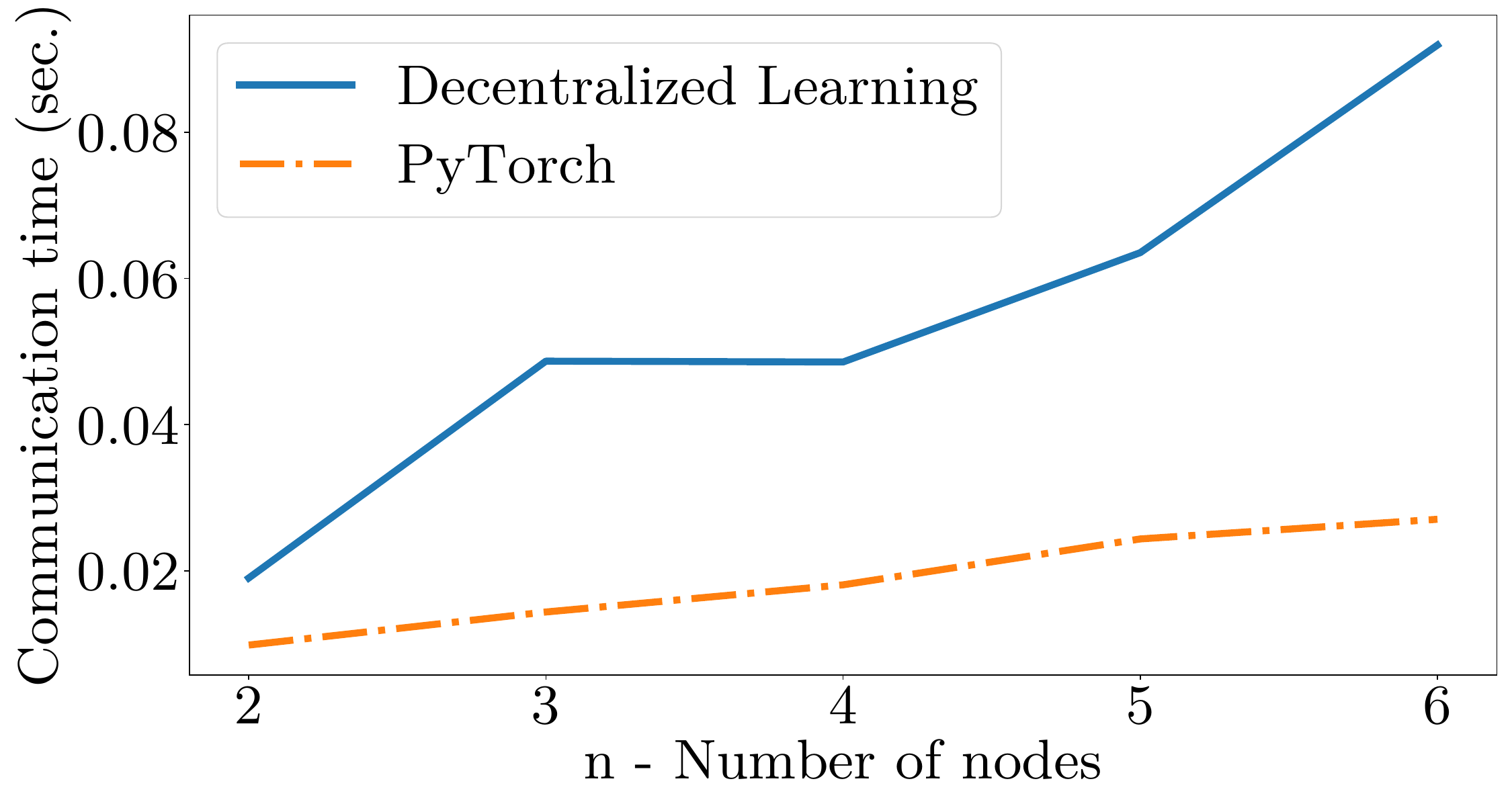}
\label{subfig:rpc-bench-n}}
\subfloat[Input dimension]{\includegraphics[width=0.49\linewidth,keepaspectratio]{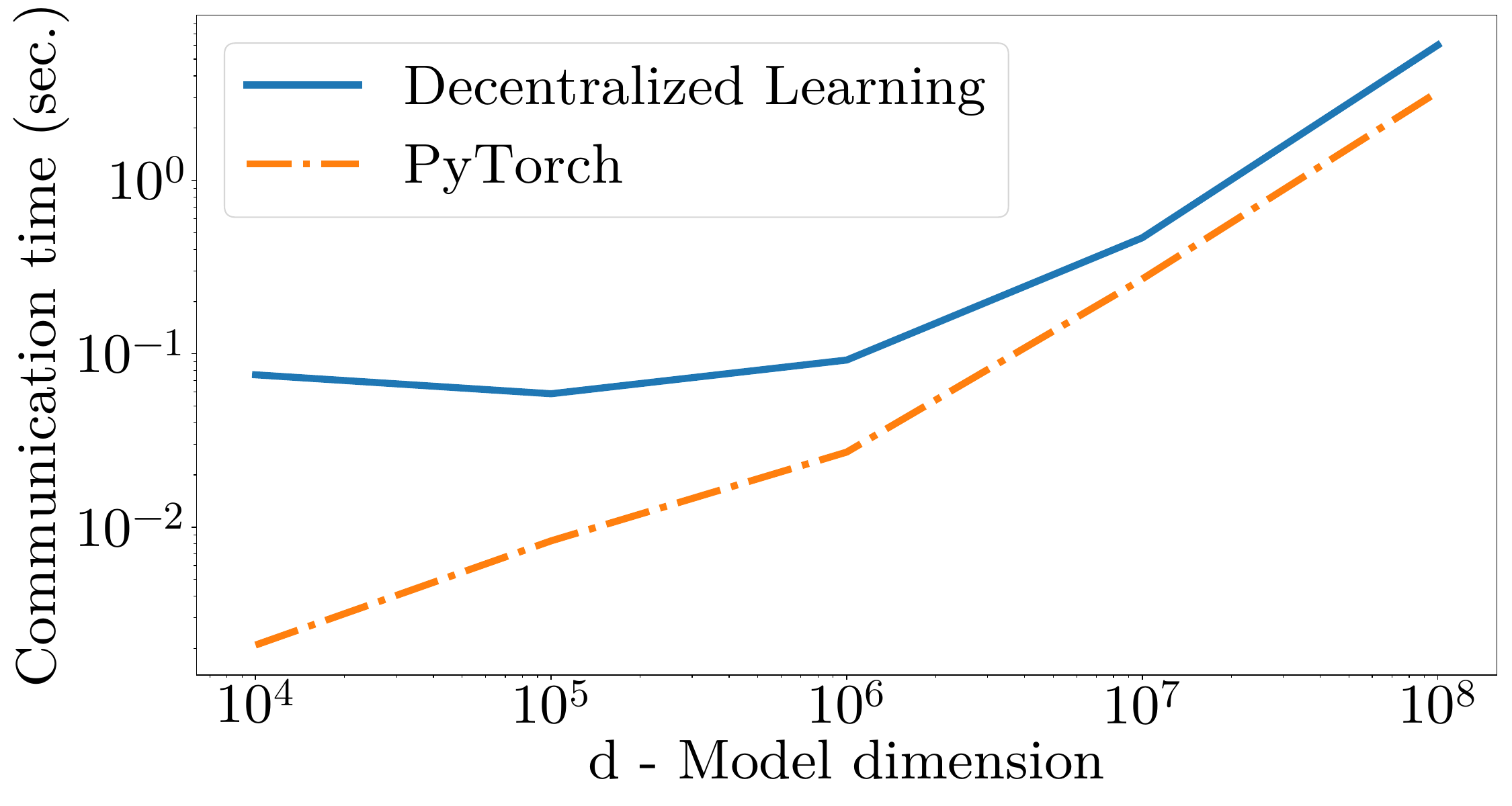}
\label{subfig:rpc-bench-d}} 
\caption{Communication time of \learn{} and vanilla baseline (deployed on GPUs) with $n$ and $d$.}
\label{fig:rpc-bench}
\vspace{-2mm}
\end{figure}

To understand why \emph{\learn} does not scale, we focus on its communication overhead with different number of nodes ($n$) and model dimension ($d$). 
Figure~\ref{fig:rpc-bench} shows the communication latency of \learn{} and vanilla baseline (PyTorch in this experiment) with different values of $n$ (with $d=10^6$) and $d$ (with $n=6$).
It is evident that increasing $d$ saturates the bandwidth quickly and increase the communication time linearly for both systems (Figure~\ref{subfig:rpc-bench-d}).
Yet, the scalability issue of \learn{} appears in Figure~\ref{subfig:rpc-bench-n}, where 
the communication time increases drastically (i.e.,\ quadratically) for \learn{} while only linearly with the vanilla competitor.
Basically, \learn{} requires $\mathcal{O}(n^2)$ messages per round while the vanilla deployments require only $\mathcal{O}(n)$ messages per round.

\begin{figure}[!th]
\centering
\vspace{-4mm}
\subfloat[Number of Byzantine workers]{\includegraphics[width=0.49\linewidth,keepaspectratio]{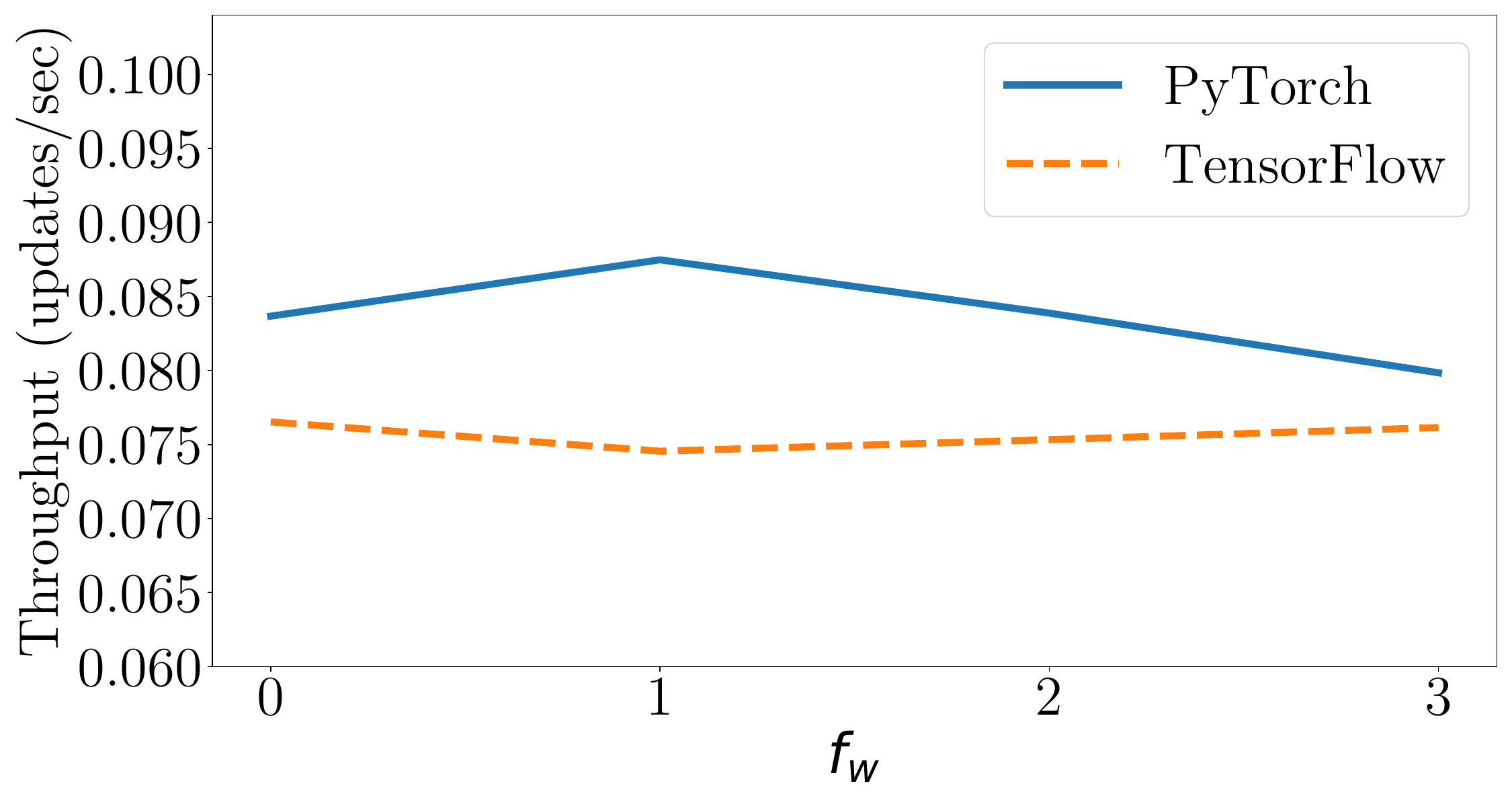}
\label{subfig:throughput-f}}
\subfloat[Number of Byzantine servers]{\includegraphics[width=0.49\linewidth,keepaspectratio]{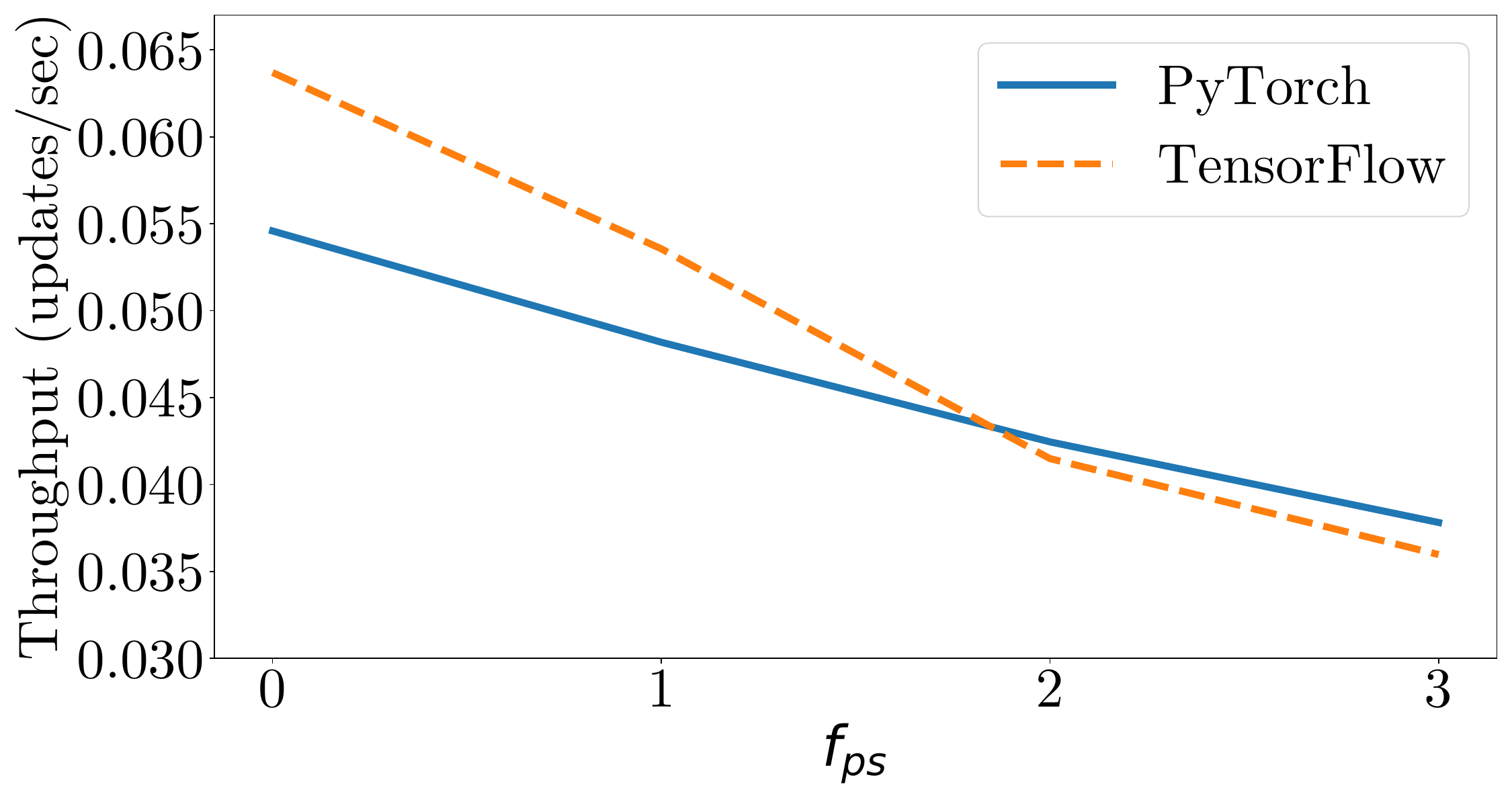}
\label{subfig:throughput-fp}} 
\caption{Throughput with increasing $f_w$ and $f_{ps}$ (on CPUs).}
\label{fig:throughput-tf}
\vspace{-4mm}
\end{figure}

\paragraph{Number of Byzantine workers}
As increasing the number of Byzantine workers ($f_w$) does not call for increasing the total number of workers, we fix $n_w$ and hence, fixing the effective batch size in all cases (in Figure~\ref{subfig:throughput-f}).
Fixing $n_w$ results in a fixed communication cost in all cases, making the throughput almost the same even with increasing $f_w$.
The same results are confirmed with our both frameworks, with a slight superiority of PyTorch to TensorFlow.

\paragraph{Number of Byzantine servers}
Increasing the number of Byzantine servers ($f_{ps}$) calls for increasing the 
number of 
server replicas ($n_{ps}$) so as to satisfy the Byzantine resilience condition: $n_{ps} \ge 3f_{ps} + 1$. 
Thus, increasing $f_{ps}$ introduces new communication links, leading to a throughput drop as shown in Figure~\ref{subfig:throughput-fp}.
Such a drop is confirmed in the \emph{state machine replication} (SMR) literature~\cite{hunt2010zookeeper,abd2005fault}, where 
the amount of drop (less than $50\%$) 
is reasonable compared to what was reported before in the literature~\cite{cowling2006hq}.
The assumption of 1 faulty parameter server introduces an overhead of $33\%$ to achieve Byzantine resilience. 
Finally we note that increasing $f_{ps}$ does not affect the number of iterations required for~convergence. 

\vspace{-0.1cm}
\section{Related Work}
\label{sec:relwork}
To the best of our knowledge, \aggregathor{}~\cite{aggregathor} is the only  implementation of a Byzantine ML system that is prior to our work.
\aggregathor{} is basically a framework 
integrated with
TensorFlow to tolerate Byzantine workers while assuming one central, trusted parameter server.
It 
relies on~\mkrum{} to robustly aggregate workers' gradients. 
\aggregathor{} relies on two components: 
the aggregation layer 
and the communication layer, which enables experimenting with lossy networks.
Though \aggregathor{} relies on the shared graph design, it disallows workers to change such a graph to combat any possible Byzantine behavior. 
The design of 
\aggregathor{} is tied to the \emph{single server, multiple workers} architecture; it cannot be used to replicate the server nor to run a peer-to-peer decentralized scheme.
Noteworthy, unlike \aggregathor{}, \system{} can be used \emph{on top of} TensorFlow or PyTorch without changing their internals nor interfaces.

On the theoretical side,
several Byzantine-resilient ML algorithms have been proposed; 
all try to mathematically bound the deviation of the aggregated gradient from the correct ones. 
Krum~\cite{krum} employs a median--like aggregation rule. 
Multi-Krum~\cite{aggregathor} generalizes the idea by averaging more gradients to benefit from additional workers.
Bulyan~\cite{bulyanPaper} addresses an attack that can trick some Byzantine--resilient algorithms by having them converge to a stable yet faulty state. 
Different variants of robust mean-based algorithms under different assumptions and scenarios were considered in \cite{xie2018generalized,yin2018byzantine}.
Kardam~\cite{kardam} uses filtering mechanisms to
achieve Byzantine resilience in an asynchronous training setup.
Zeno~\cite{xie2018zeno} and Zeno++~\cite{xie2019zeno++} achieve Byzantine resilience using a performance--based ranking approach 
in synchronous and asynchronous settings respectively.
Draco~\cite{chen2018draco} uses a coding scheme to restore correct gradients using redundant computations. 
Detox~\cite{rajput2019detox} extends this idea by combining coding schemes with robust aggregation to hit the sweet spot in the resilience--optimality spectrum. 
ByzSGD~\cite{el2020genuinely} shows how to combine robust GARs 
to tolerate Byzantine servers too.
In particular, it replicates the parameter server on multiple machines while letting them communicate to limit the divergence among their model states. 
ByRDiE~\cite{yang2019byrdie} and BRIDGE~\cite{yang2019bridge} combine both robust aggregation with performance--based ranking to achieve Byzantine resilience in the decentralized~settings.
\system{} uses robust aggregation and
already implements many of the mentioned GARs. \system{} can straightforwardly include the other ones.


 


The problem of tolerating benign (i.e.,\ crash) failures of parameter vectors was also addressed in the literature.
Qiao et al.~\cite{qiao2019fault} leverage the self--correcting behavior of SGD to tolerate such failures.
Other proposals addressed the problem of making the parameter server crash-resilient~\cite{li2013parameter,chilimbi2014project} using Paxos~\cite{lamport2001paxos}.  
Others rely on checkpoints or live replication~\cite{abadi2016tensorflow} of the parameter server.
However, we believe that extending those tools to the Byzantine context would be prohibitive.
\vspace{-0.1cm}
\section{Concluding Remarks}
\label{sec:conc}

This paper presents \system{}, a library to build Byzantine machine learning (ML) applications on top of popular frameworks such as TensorFlow and PyTorch, while achieving \emph{transparency}: applications developed with either framework do not need to change their interfaces to be made Byzantine resilient. 
\system{} supports multiple statistically--robust gradient aggregation rules (GARs), which can be combined in various ways for different resilience properties.
In some situations, GARs fail to ensure Byzantine resilience when the underlying assumption on a  bounded variance  is not satisfied~\cite{baruch2019little}.
Yet, several techniques were proposed for variance reduction, e.g.,\ ~\cite{wang2013variance,allen2016variance,park2019accelerated}, which help restore the resilience guarantees of such GARs~\cite{el2020distributed}. 
Such techniques can be added seamlessly to \system{} without affecting its throughput performance as they basically only change the optimization function.
In the same vein, we believe \system{} could be also used to implement applications that combine privacy and security properties with
Byzantine resilience as in~\cite{he2020secure,munoz2019byzantine}.
Our code is available~\cite{code} and will be open-sourced upon publication.
Our evaluation of \system{} (using three Byzantine ML applications) showed that Byzantine resilience, unlike crash resilience, induces an inherent loss in the final accuracy and that the throughput overhead of Byzantine resilience is moderate compared to crash resilience.
Furthermore, we showed that (1) the Byzantine resilience overhead comes more from communication than from aggregation, and that (2) the overhead of tolerating Byzantine servers is much more than that of tolerating Byzantine workers.

\section*{Acknowledgment}
This work has been supported in part by the Swiss National
Science Foundation (FNS grant 200021\_182542/1).

Experiments presented in this paper were carried out using the Grid’5000 testbed, supported by a scientific interest group hosted by Inria and including CNRS, RENATER
and several Universities as well as other organizations (see \url{https://www.grid5000.fr}).

\bibliographystyle{plain}
\bibliography{bib}

\begin{thebibliography}{10}

\bibitem{aggregathorcode}
Aggregathor source code.
\newblock \url{https://github.com/LPD-EPFL/AggregaThor}.

\bibitem{cifar}
Cifar dataset.
\newblock \url{https://www.cs.toronto.edu/~kriz/cifar.html}.

\bibitem{g5k}
Grid5000.
\newblock \url{https://www.grid5000.fr/}.

\bibitem{mnist}
Mnist dataset.
\newblock {\url{http://yann.lecun.com/exdb/mnist/}}.

\bibitem{code}
\system{} source code.
\newblock \url{https://github.com/LPD-EPFL/Garfield}.

\bibitem{abadi2016tensorflow}
Mart{\'\i}n Abadi, Paul Barham, Jianmin Chen, Zhifeng Chen, Andy Davis, Jeffrey
  Dean, Matthieu Devin, Sanjay Ghemawat, Geoffrey Irving, Michael Isard, et~al.
\newblock Tensorflow: A system for large-scale machine learning.
\newblock In {\em OSDI}, 2016.

\bibitem{abd2005fault}
Michael Abd-El-Malek, Gregory~R Ganger, Garth~R Goodson, Michael~K Reiter, and
  Jay~J Wylie.
\newblock Fault-scalable byzantine fault-tolerant services.
\newblock {\em ACM SIGOPS Operating Systems Review}, 39:59--74, 2005.

\bibitem{alistarh2018byzantine}
Dan Alistarh, Zeyuan Allen-Zhu, and Jerry Li.
\newblock {B}yzantine stochastic gradient descent.
\newblock In {\em Neural Information Processing Systems, to appear}, 2018.

\bibitem{allen2016variance}
Zeyuan Allen-Zhu and Elad Hazan.
\newblock Variance reduction for faster non-convex optimization.
\newblock In {\em International conference on machine learning}, pages
  699--707, 2016.

\bibitem{baruch2019little}
Moran Baruch, Gilad Baruch, and Yoav Goldberg.
\newblock A little is enough: Circumventing defenses for distributed learning.
\newblock {\em arXiv preprint arXiv:1902.06156}, 2019.

\bibitem{krum}
Peva Blanchard, El~Mahdi El~Mhamdi, Rachid Guerraoui, and Julien Stainer.
\newblock Machine learning with adversaries: {B}yzantine tolerant gradient
  descent.
\newblock In {\em Neural Information Processing Systems}, pages 118--128, 2017.

\bibitem{bloom2017self}
Cara Bloom, Joshua Tan, Javed Ramjohn, and Lujo Bauer.
\newblock Self-driving cars and data collection: Privacy perceptions of
  networked autonomous vehicles.
\newblock In {\em Thirteenth Symposium on Usable Privacy and Security
  ($\{$SOUPS$\}$ 2017)}, pages 357--375, 2017.

\bibitem{bonawitz2017practical}
Keith Bonawitz, Vladimir Ivanov, Ben Kreuter, Antonio Marcedone, H~Brendan
  McMahan, Sarvar Patel, Daniel Ramage, Aaron Segal, and Karn Seth.
\newblock Practical secure aggregation for privacy-preserving machine learning.
\newblock In {\em Proceedings of the 2017 ACM SIGSAC Conference on Computer and
  Communications Security}, pages 1175--1191. ACM, 2017.

\bibitem{chen2018draco}
Lingjiao Chen, Hongyi Wang, Zachary Charles, and Dimitris Papailiopoulos.
\newblock Draco: Byzantine-resilient distributed training via redundant
  gradients.
\newblock In {\em International Conference on Machine Learning}, pages
  902--911, 2018.

\bibitem{chen2017distributed}
Yudong Chen, Lili Su, and Jiaming Xu.
\newblock Distributed statistical machine learning in adversarial settings:
  {B}yzantine gradient descent.
\newblock {\em arXiv preprint arXiv:1705.05491}, 2017.

\bibitem{chilimbi2014project}
Trishul~M Chilimbi, Yutaka Suzue, Johnson Apacible, and Karthik Kalyanaraman.
\newblock Project adam: Building an efficient and scalable deep learning
  training system.
\newblock In {\em OSDI}, volume~14, pages 571--582, 2014.

\bibitem{cowling2006hq}
James Cowling, Daniel Myers, Barbara Liskov, Rodrigo Rodrigues, and Liuba
  Shrira.
\newblock Hq replication: A hybrid quorum protocol for byzantine fault
  tolerance.
\newblock In {\em Proceedings of the 7th symposium on Operating systems design
  and implementation}, pages 177--190. USENIX Association, 2006.

\bibitem{aggregathor}
Georgios Damaskinos, El~Mahdi El~Mhamdi, Rachid Guerraoui, Arsany Guirguis, and
  S{\'e}bastien Rouault.
\newblock Aggregathor: Byzantine machine learning via robust gradient
  aggregation.
\newblock In {\em SysML}, 2019.

\bibitem{kardam}
Georgios Damaskinos, El~Mahdi El~Mhamdi, Rachid Guerraoui, Rhicheek Patra,
  Mahsa Taziki, et~al.
\newblock Asynchronous byzantine machine learning (the case of sgd).
\newblock In {\em ICML}, pages 1153--1162, 2018.

\bibitem{el2020collaborative}
El-Mahdi El-Mhamdi, Rachid Guerraoui, Arsany Guirguis, L{\^e}~Nguy{\^e}n Hoang,
  and S{\'e}bastien Rouault.
\newblock Collaborative learning as an agreement problem.
\newblock {\em arXiv preprint arXiv:2008.00742}, 2020.

\bibitem{el2020genuinely}
El-Mahdi El-Mhamdi, Rachid Guerraoui, Arsany Guirguis, L{\^e}~Nguy{\^e}n Hoang,
  and S{\'e}bastien Rouault.
\newblock Genuinely distributed byzantine machine learning.
\newblock In {\em Proceedings of the 39th Symposium on Principles of
  Distributed Computing}, pages 355--364, 2020.

\bibitem{bulyanPaper}
El~Mahdi El~Mhamdi, Rachid Guerraoui, and S{\'e}bastien Rouault.
\newblock The hidden vulnerability of distributed learning in {B}yzantium.
\newblock In Jennifer Dy and Andreas Krause, editors, {\em Proceedings of the
  35th International Conference on Machine Learning}, volume~80 of {\em
  Proceedings of Machine Learning Research}, pages 3521--3530,
  Stockholmsmässan, Stockholm Sweden, 10--15 Jul 2018. PMLR.

\bibitem{el2020distributed}
El-Mahdi El-Mhamdi, Rachid Guerraoui, and S{\'e}bastien Rouault.
\newblock Distributed momentum for byzantine-resilient learning.
\newblock {\em arXiv preprint arXiv:2003.00010}, 2020.

\bibitem{esteva2017dermatologist}
Andre Esteva, Brett Kuprel, Roberto~A Novoa, Justin Ko, Susan~M Swetter,
  Helen~M Blau, and Sebastian Thrun.
\newblock Dermatologist-level classification of skin cancer with deep neural
  networks.
\newblock {\em Nature}, 542(7639):115, 2017.

\bibitem{ghiassi2019robust}
Amirmasoud Ghiassi, Taraneh Younesian, Zhilong Zhao, Robert Birke, Valerio
  Schiavoni, and Lydia~Y Chen.
\newblock Robust (deep) learning framework against dirty labels and beyond.
\newblock In {\em 2019 First IEEE International Conference on Trust, Privacy
  and Security in Intelligent Systems and Applications (TPS-ISA)}, pages
  236--244. IEEE, 2019.

\bibitem{DBLP:journals/corr/HeZR016}
Kaiming He, Xiangyu Zhang, Shaoqing Ren, and Jian Sun.
\newblock Identity mappings in deep residual networks.
\newblock {\em CoRR}, abs/1603.05027, 2016.

\bibitem{colalearning}
Lie He, An~Bian, and Martin Jaggi.
\newblock Cola: Decentralized linear learning.
\newblock In S.~Bengio, H.~Wallach, H.~Larochelle, K.~Grauman, N.~Cesa-Bianchi,
  and R.~Garnett, editors, {\em Advances in Neural Information Processing
  Systems 31}, pages 4536--4546. Curran Associates, Inc., 2018.

\bibitem{he2020secure}
Lie He, Sai~Praneeth Karimireddy, and Martin Jaggi.
\newblock Secure byzantine-robust machine learning.
\newblock {\em arXiv preprint arXiv:2006.04747}, 2020.

\bibitem{hunt2010zookeeper}
Patrick Hunt, Mahadev Konar, Flavio~Paiva Junqueira, and Benjamin Reed.
\newblock Zookeeper: Wait-free coordination for internet-scale systems.
\newblock In {\em USENIX annual technical conference}, volume~8. Boston, MA,
  USA, 2010.

\bibitem{branchless-median}
M.~{Kachelrieß}.
\newblock Branchless vectorized median filtering.
\newblock In {\em 2009 IEEE Nuclear Science Symposium Conference Record
  (NSS/MIC)}, pages 4099--4105, Oct 2009.

\bibitem{kim1many}
Larry Kim.
\newblock How many ads does google serve in a day?
\newblock {\em URL http://goo. gl/oIidXO. http://goo. gl/oIidXO}, 1(1), 2012.

\bibitem{lamport2001paxos}
Leslie Lamport et~al.
\newblock Paxos made simple.
\newblock {\em ACM Sigact News}, 32(4):18--25, 2001.

\bibitem{lamport1982Byzantine}
Leslie Lamport, Robert Shostak, and Marshall Pease.
\newblock The {B}yzantine generals problem.
\newblock {\em TOPLAS}, 4(3):382--401, 1982.

\bibitem{li2019rsa}
Liping Li, Wei Xu, Tianyi Chen, Georgios~B Giannakis, and Qing Ling.
\newblock Rsa: Byzantine-robust stochastic aggregation methods for distributed
  learning from heterogeneous datasets.
\newblock In {\em Proceedings of the AAAI Conference on Artificial
  Intelligence}, volume~33, pages 1544--1551, 2019.

\bibitem{li2014scaling}
Mu~Li, David~G Andersen, Jun~Woo Park, Alexander~J Smola, Amr Ahmed, Vanja
  Josifovski, James Long, Eugene~J Shekita, and Bor-Yiing Su.
\newblock Scaling distributed machine learning with the parameter server.
\newblock In {\em OSDI}, volume~1, page~3, 2014.

\bibitem{li2013parameter}
Mu~Li, Li~Zhou, Zichao Yang, Aaron Li, Fei Xia, David~G Andersen, and Alexander
  Smola.
\newblock Parameter server for distributed machine learning.
\newblock In {\em Big Learning NIPS Workshop}, volume~6, page~2, 2013.

\bibitem{mcmahan2013ad}
H~Brendan McMahan, Gary Holt, David Sculley, Michael Young, Dietmar Ebner,
  Julian Grady, Lan Nie, Todd Phillips, Eugene Davydov, Daniel Golovin, et~al.
\newblock Ad click prediction: a view from the trenches.
\newblock In {\em Proceedings of the 19th ACM SIGKDD international conference
  on Knowledge discovery and data mining}, pages 1222--1230. ACM, 2013.

\bibitem{meng2016mllib}
Xiangrui Meng, Joseph Bradley, Burak Yavuz, Evan Sparks, Shivaram Venkataraman,
  Davies Liu, Jeremy Freeman, DB~Tsai, Manish Amde, Sean Owen, et~al.
\newblock Mllib: Machine learning in apache spark.
\newblock {\em JMLR}, 17(1):1235--1241, 2016.

\bibitem{munoz2019byzantine}
Luis Mu{\~n}oz-Gonz{\'a}lez, Kenneth~T Co, and Emil~C Lupu.
\newblock Byzantine-robust federated machine learning through adaptive model
  averaging.
\newblock {\em arXiv preprint arXiv:1909.05125}, 2019.

\bibitem{musser1997introspective}
David~R Musser.
\newblock Introspective sorting and selection algorithms.
\newblock {\em Software: Practice and Experience}, 27(8):983--993, 1997.

\bibitem{park2019accelerated}
Jay~H Park, Sunghwan Kim, Jinwon Lee, Myeongjae Jeon, and Sam~H Noh.
\newblock Accelerated training for cnn distributed deep learning through
  automatic resource-aware layer placement.
\newblock {\em arXiv preprint arXiv:1901.05803}, 2019.

\bibitem{paszke2019pytorch}
Adam Paszke, Sam Gross, Francisco Massa, Adam Lerer, James Bradbury, Gregory
  Chanan, Trevor Killeen, Zeming Lin, Natalia Gimelshein, Luca Antiga, et~al.
\newblock Pytorch: An imperative style, high-performance deep learning library.
\newblock In {\em Advances in neural information processing systems}, pages
  8026--8037, 2019.

\bibitem{patarasuk2009bandwidth}
Pitch Patarasuk and Xin Yuan.
\newblock Bandwidth optimal all-reduce algorithms for clusters of workstations.
\newblock {\em Journal of Parallel and Distributed Computing}, 69(2):117--124,
  2009.

\bibitem{qiao2019fault}
Aurick Qiao, Bryon Aragam, Bingjing Zhang, and Eric Xing.
\newblock Fault tolerance in iterative-convergent machine learning.
\newblock In {\em International Conference on Machine Learning}, pages
  5220--5230, 2019.

\bibitem{rajput2019detox}
Shashank Rajput, Hongyi Wang, Zachary Charles, and Dimitris Papailiopoulos.
\newblock Detox: A redundancy-based framework for faster and more robust
  gradient aggregation.
\newblock {\em arXiv preprint arXiv:1907.12205}, 2019.

\bibitem{rao2018deep}
Qing Rao and Jelena Frtunikj.
\newblock Deep learning for self-driving cars: chances and challenges.
\newblock In {\em 2018 IEEE/ACM 1st International Workshop on Software
  Engineering for AI in Autonomous Systems (SEFAIAS)}, pages 35--38. IEEE,
  2018.

\bibitem{rousseeuw1985multivariate}
Peter~J Rousseeuw.
\newblock Multivariate estimation with high breakdown point.
\newblock {\em Mathematical statistics and applications}, 8:283--297, 1985.

\bibitem{rumelhart1986learning}
David~E Rumelhart, Geoffrey~E Hinton, and Ronald~J Williams.
\newblock Learning representations by back-propagating errors.
\newblock {\em nature}, 323(6088):533--536, 1986.

\bibitem{shallue2018measuring}
Christopher~J Shallue, Jaehoon Lee, Joe Antognini, Jascha Sohl-Dickstein, Roy
  Frostig, and George~E Dahl.
\newblock Measuring the effects of data parallelism on neural network training.
\newblock {\em arXiv preprint arXiv:1811.03600}, 2018.

\bibitem{simonyan2014very}
Karen Simonyan and Andrew Zisserman.
\newblock Very deep convolutional networks for large-scale image recognition.
\newblock {\em arXiv preprint arXiv:1409.1556}, 2014.

\bibitem{vanhaesebrouck2016decentralized}
Paul Vanhaesebrouck, Aur{\'e}lien Bellet, and Marc Tommasi.
\newblock Decentralized collaborative learning of personalized models over
  networks.
\newblock In {\em AISTATS}, 2017.

\bibitem{protobuf}
Kenton Varda.
\newblock Protocol buffers.
\newblock \url{https://github.com/protocolbuffers/protobuf}.

\bibitem{lilisu2019}
Pooja Vyavahare, Lili Su, and Nitin~H Vaidya.
\newblock Distributed learning with adversarial agents under relaxed network
  condition.
\newblock {\em arXiv preprint arXiv:1901.01943}, 2019.

\bibitem{wang2013variance}
Chong Wang, Xi~Chen, Alexander~J Smola, and Eric~P Xing.
\newblock Variance reduction for stochastic gradient optimization.
\newblock {\em Advances in Neural Information Processing Systems}, 26:181--189,
  2013.

\bibitem{xie2018generalized}
Cong Xie, Oluwasanmi Koyejo, and Indranil Gupta.
\newblock Generalized {B}yzantine-tolerant sgd.
\newblock {\em arXiv preprint arXiv:1802.10116}, 2018.

\bibitem{xie2018zeno}
Cong Xie, Oluwasanmi Koyejo, and Indranil Gupta.
\newblock Zeno: Byzantine-suspicious stochastic gradient descent.
\newblock {\em arXiv preprint arXiv:1805.10032}, 2018.

\bibitem{xie2018faster}
Cong Xie, Oluwasanmi~O Koyejo, and Indranil Gupta.
\newblock Faster distributed synchronous sgd with weak synchronization.
\newblock 2018.

\bibitem{xie2019fall}
Cong Xie, Sanmi Koyejo, and Indranil Gupta.
\newblock Fall of empires: Breaking byzantine-tolerant sgd by inner product
  manipulation.
\newblock {\em arXiv preprint arXiv:1903.03936}, 2019.

\bibitem{xie2019zeno++}
Cong Xie, Sanmi Koyejo, and Indranil Gupta.
\newblock Zeno++: Robust fully asynchronous sgd.
\newblock {\em arXiv preprint arXiv:1903.07020}, 2019.

\bibitem{yang2019bridge}
Zhixiong Yang and Waheed~U Bajwa.
\newblock Bridge: Byzantine-resilient decentralized gradient descent.
\newblock {\em arXiv preprint arXiv:1908.08098}, 2019.

\bibitem{yang2019byrdie}
Zhixiong Yang and Waheed~U Bajwa.
\newblock Byrdie: Byzantine-resilient distributed coordinate descent for
  decentralized learning.
\newblock {\em IEEE Transactions on Signal and Information Processing over
  Networks}, 5(4):611--627, 2019.

\bibitem{yin2018byzantine}
Dong Yin, Yudong Chen, Kannan Ramchandran, and Peter Bartlett.
\newblock {B}yzantine-robust distributed learning: Towards optimal statistical
  rates.
\newblock {\em arXiv preprint arXiv:1803.01498}, 2018.

\end{thebibliography}

\newpage
\setcounter{section}{0}
\vspace{0.5cm}
\begin{center}
\vspace{1cm}
\hrule\vspace{1cm}
{\Large Additional Experiments}
\vspace{1cm}\hrule{}\vspace{1.5cm}
\end{center}
\vspace{1cm}

\begin{figure*}[!ht]
\centering 
\vspace{-4mm}
\subfloat[Convergence with CifarNet]{\includegraphics[width=0.49\linewidth,keepaspectratio]{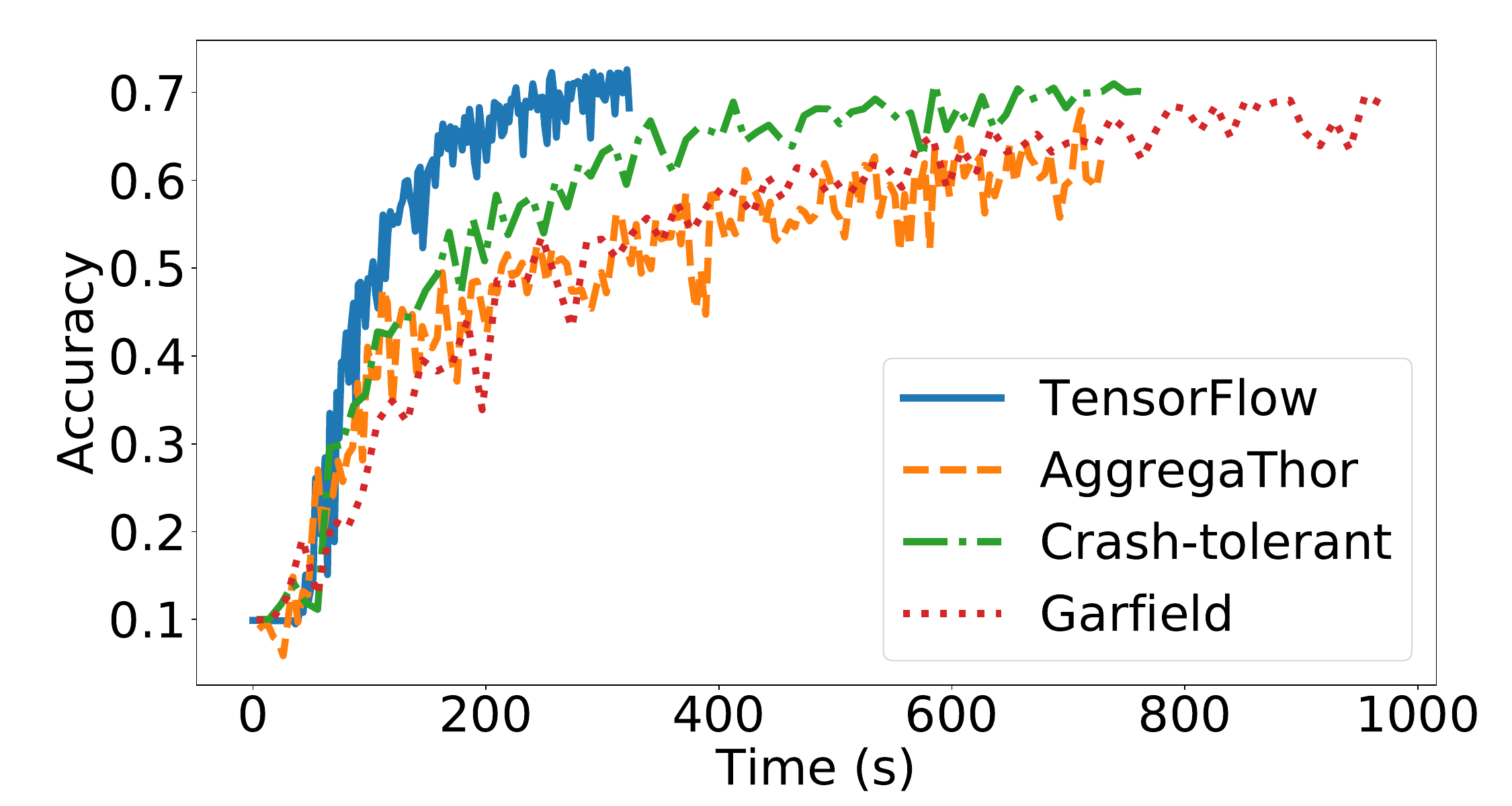}
\label{subfig:conv_tf1}}
\subfloat[Convergence with ResNet50]{\includegraphics[width=0.49\linewidth,keepaspectratio]{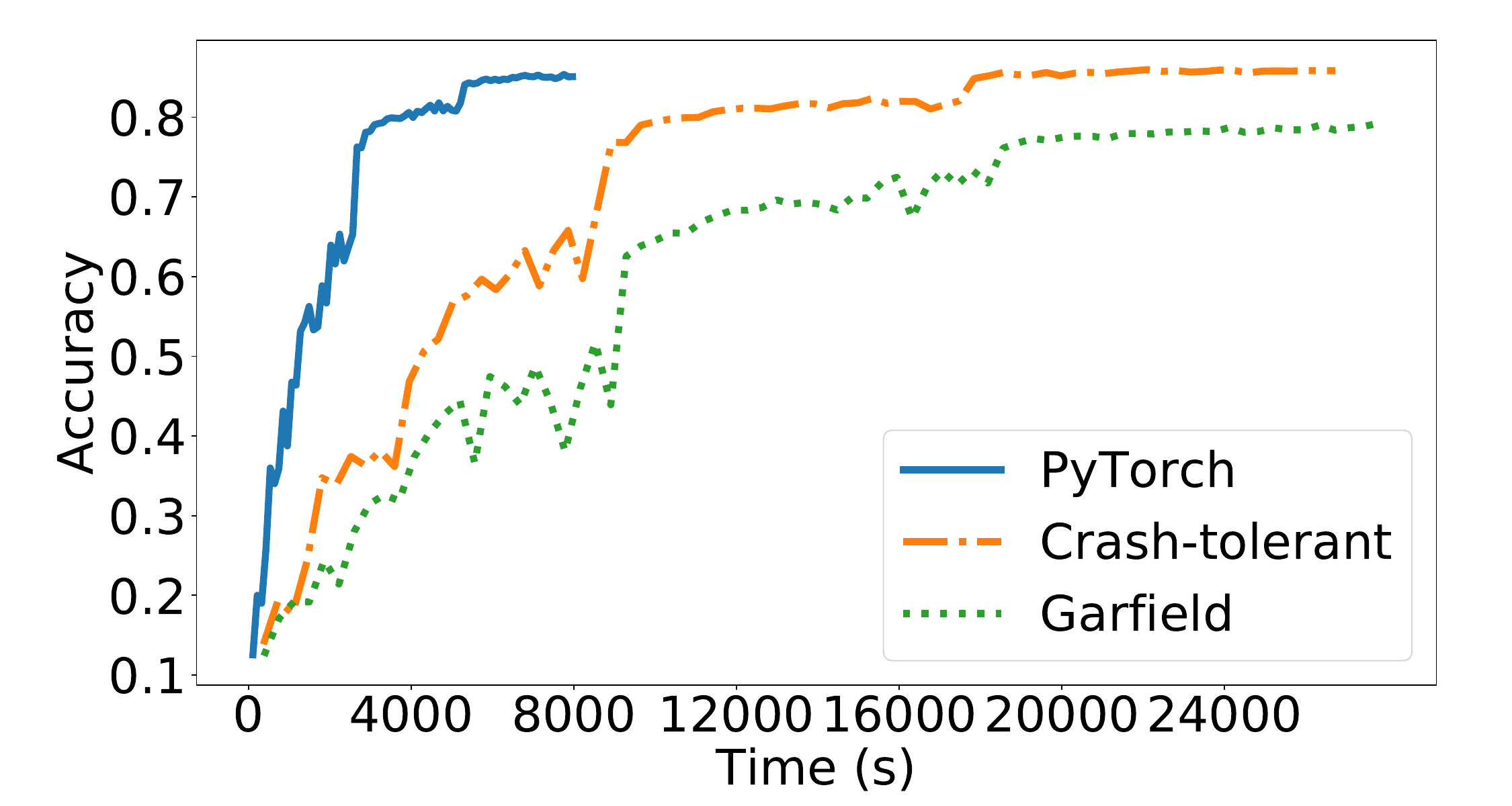}
\label{subfig:conv_pt1}}
\vspace{-2mm}
\caption{Convergence of \system{} over time with respect to other baselines using two models.}
\vspace{-4mm}
\label{fig:conv}
\end{figure*}

\section{Convergence with Time}
As discussed in the main manuscript, the end metric that one would use to assess the training progress is convergence over time. Such a metric combines both metrics presented in the main manuscript: convergence per epoch and throughput. Figure~\ref{fig:conv} is the mirror of Figure~4 in the main manuscript, which shows the convergence of \system{} compared to the baselines with the training rounds. Figure~\ref{fig:conv} shows the results of the very same experiments, yet with time rather than training rounds. Confirming the results we give concerning the throughput of the compared systems, Figure~\ref{subfig:conv_tf1} shows that the vanilla deployment, i.e.,\ TensorFlow converges faster than the crash--tolerant protocol, which converges faster than the Byzantine--resilient protocols. In addition, AggregaThor is faster than \system{}. 
Figure~\ref{subfig:conv_pt1} sheds even more light on the cost of fault--tolerance in general. For instance, although Figure~4b (in the main manuscript) shows that the crash--tolerant protocol can reach the same final accuracy as the vanilla deployment, i.e.,\ PyTorch, Figure~\ref{subfig:conv_pt1} shows that such an accuracy would be reached (by the crash-tolerant protocol) in more than 3x time than the vanilla deployment. Such a figure shows that the cost of tolerating mere crash failures is not negligible. Furthermore, it emphasizes the accuracy loss introduced by the Byzantine--resilient deployment, i.e.,\ \system{}. Finally, we can observe from this figure that the cost of Byzantine--resilience, compared to crash--resilience, is not big, in terms of throughput.

\begin{figure*}[!ht]
\centering 
\vspace{-4mm}
\subfloat[Convergence with number of iterations]{\includegraphics[width=0.49\linewidth,keepaspectratio]{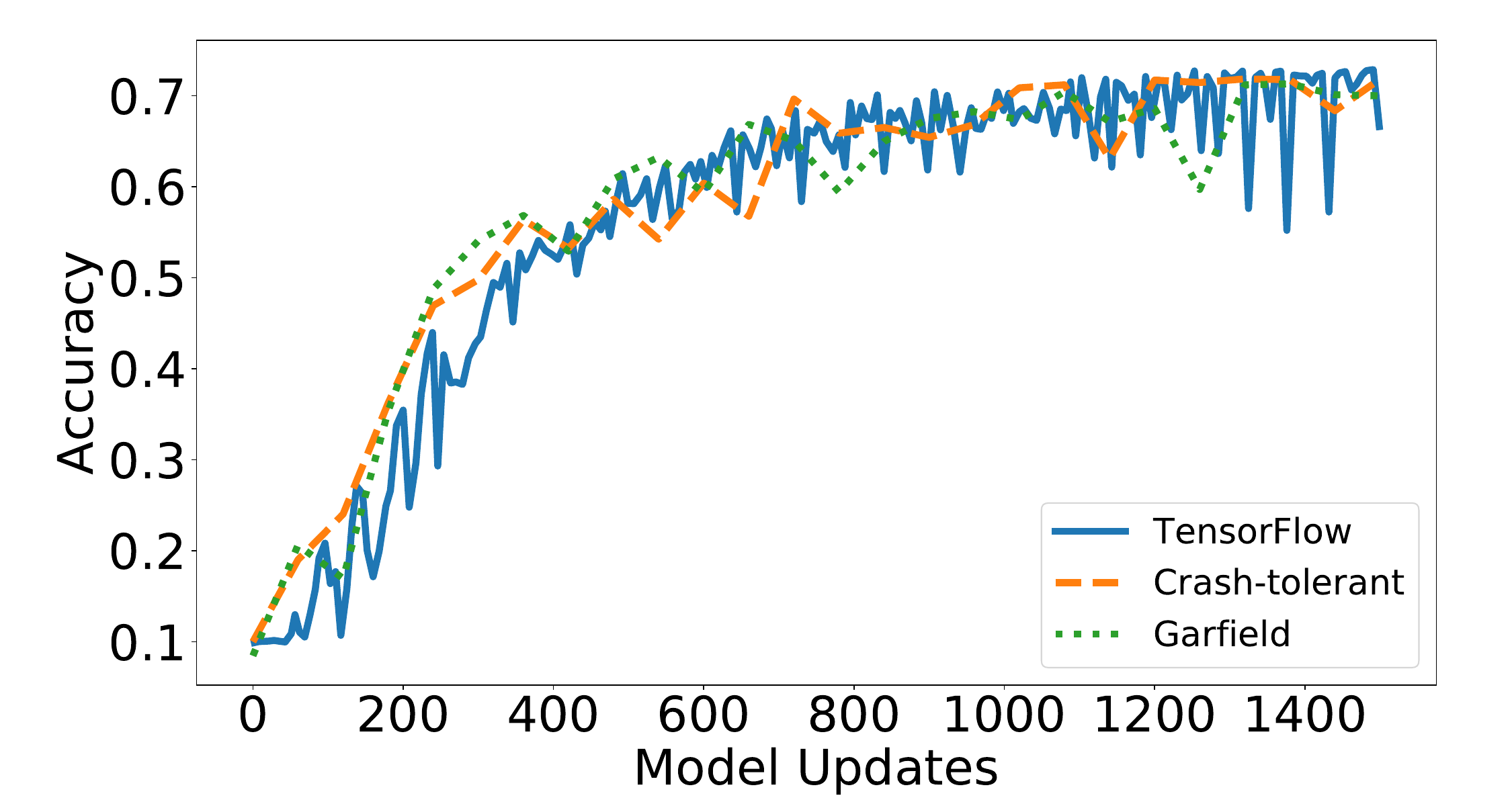}
\label{subfig:epoch}}
\subfloat[Convergence with time]{\includegraphics[width=0.49\linewidth,keepaspectratio]{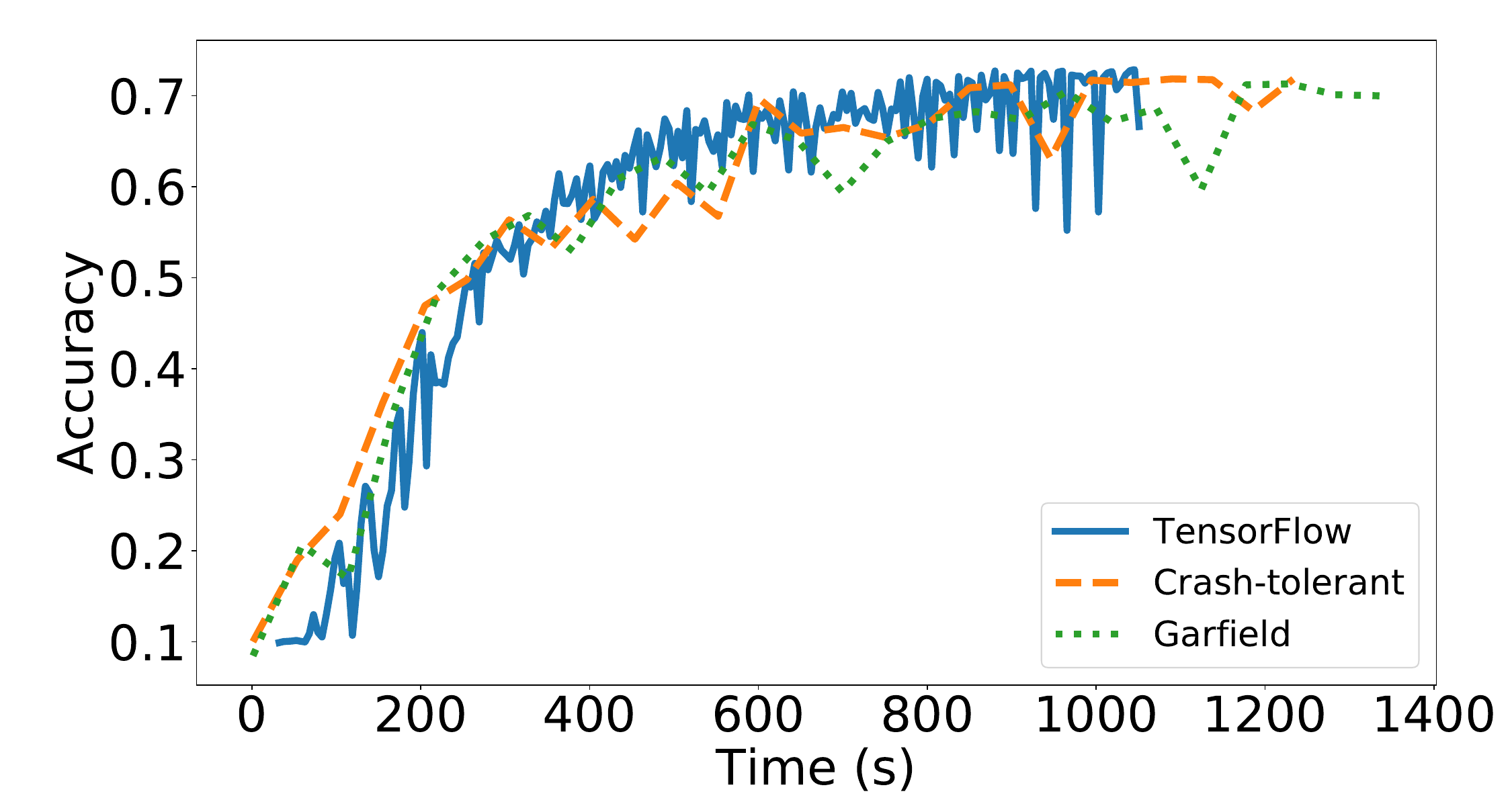}
\label{subfig:time}}
\vspace{-2mm}
\caption{Convergence of \system{}'s protocol while using MDA as a GAR.}
\vspace{-4mm}
\label{fig:conv_mda}
\end{figure*}

\section{Convergence with MDA as a GAR}
In this section, we show the convergence of \system{}'s protocol while using MDA as a GAR. We use the CPU--based cluster and the TensorFlow--based implementation in this experiment. We use the same setup as in Section 5.2 (with the default setup of TensorFlow--based experiments) in the main paper.

Figure~\ref{fig:conv_mda}\protect\subref{subfig:epoch} shows that all compared systems achieve almost the same convergence rate. Moreover, it shows that deploying the Byzantine variant of \system{} does not add any overhead compared to the correct one (vanilla TensorFlow), in terms of convergence steps. However, the cost of resilience appears in Figure~\ref{fig:conv_mda}\protect\subref{subfig:time} which depicts the convergence rate against time. For example, vanilla TensorFlow reaches accuracy of $60\%$ in 364 seconds, which is $15\%$ better than the crash--tolerant deployment, where the Byzantine deployment reaches the same accuracy level in $23\%$ more time (which is the cost of Byzantine--resilience) compared to the latter.

\section{Effect of The Number of Byzantine Machines}
In this section, we show the effect of changing the number of both Byzantine workers and Byzantine servers on the performance of \system{}. Our metric here is throughput and we use the same setup as the throughput experiments in the main manuscript. We run experiments on both CPUs (using TensorFlow) and GPUs (using PyTorch).

\begin{figure}[!ht]
\centering
\vspace{-2mm}
\subfloat[CPU]{\includegraphics[width=0.49\linewidth,keepaspectratio]{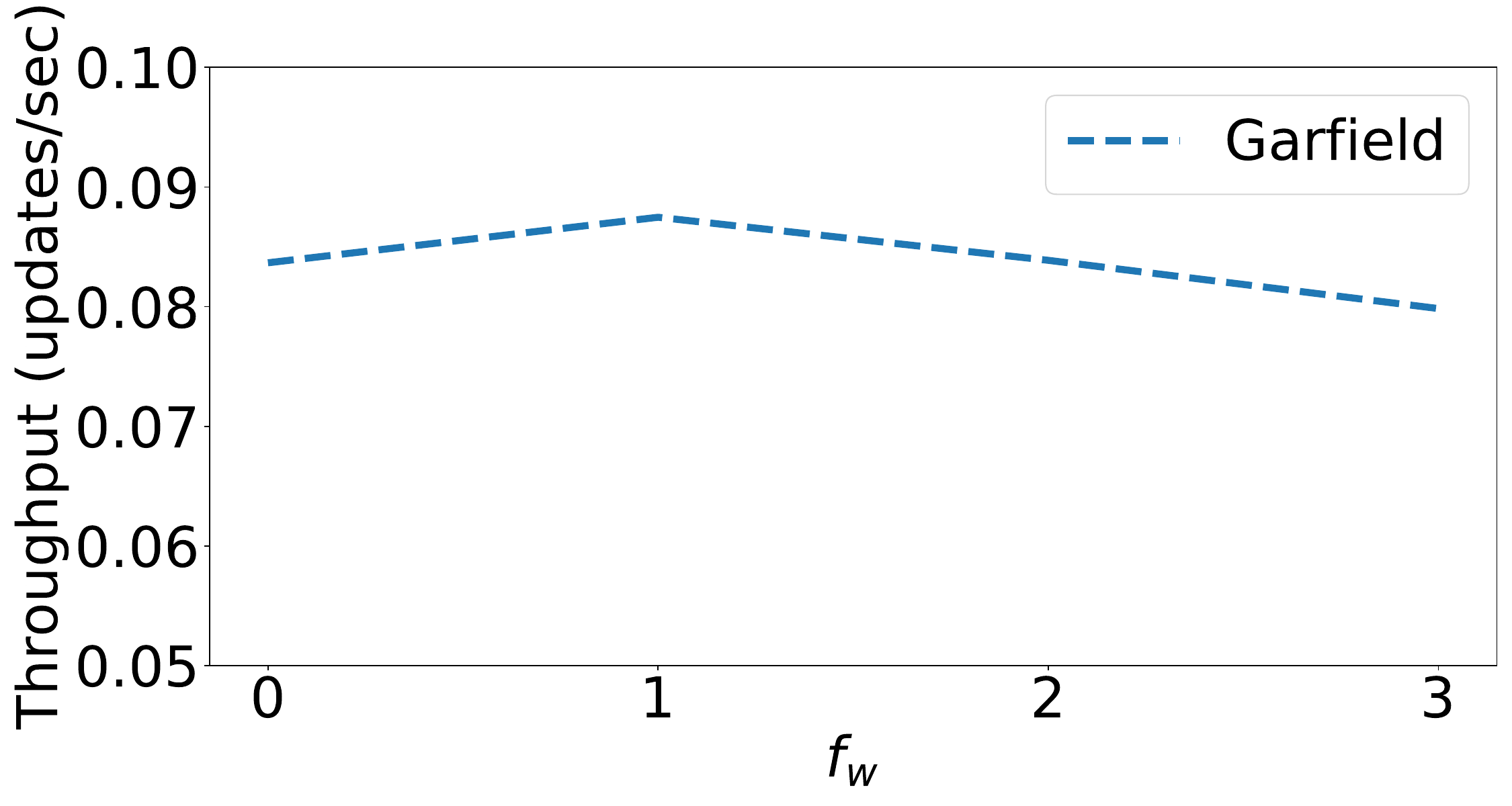}
\label{subfig:throughput-f-tf}}
\subfloat[GPU]{\includegraphics[width=0.49\linewidth,keepaspectratio]{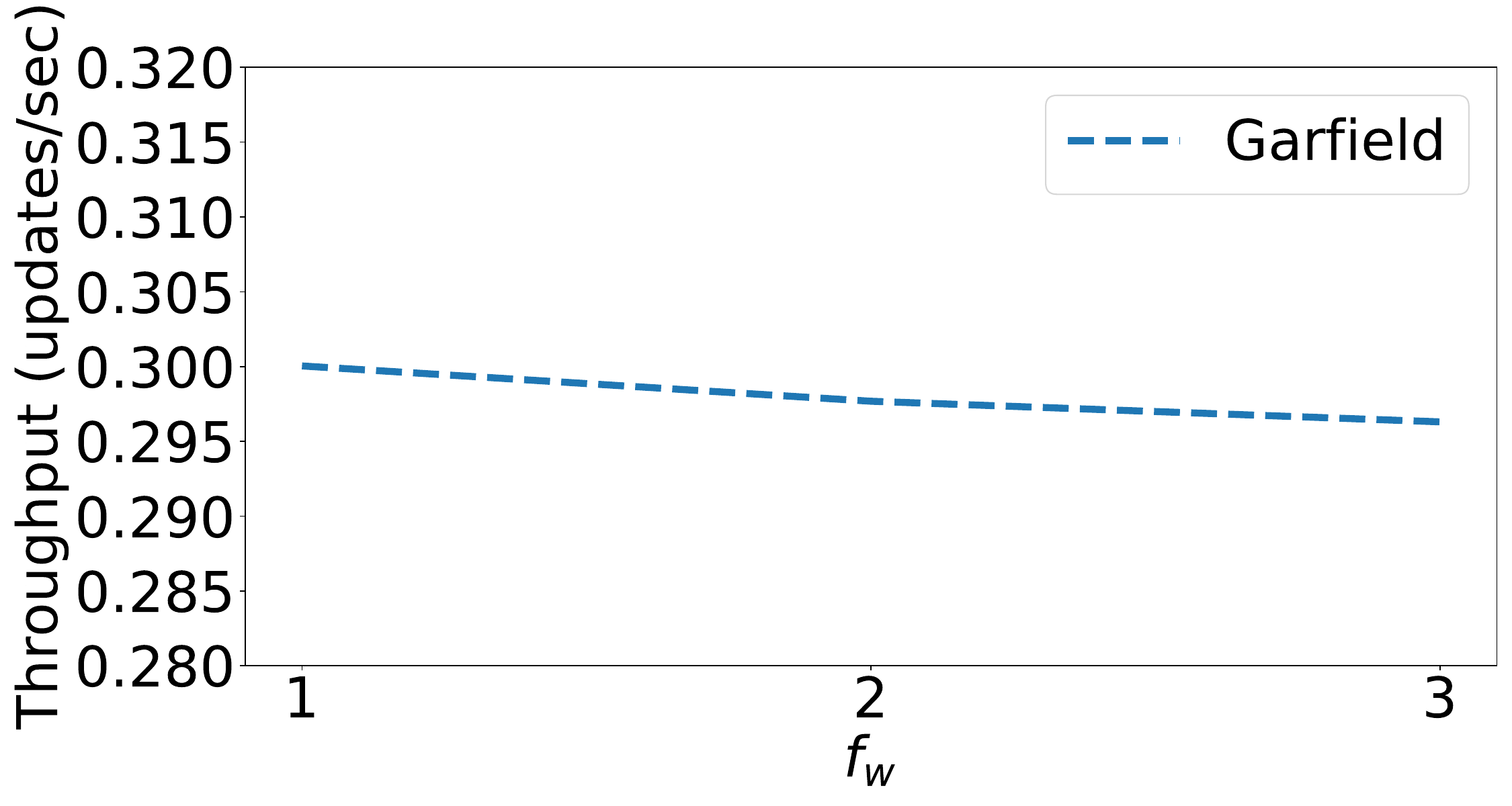}
\label{subfig:throughput-f-pt}} 
\vspace{-2mm}
\caption{Throughput of \system{} with different number of Byzantine workers.}
\label{fig:throughput-f}
\vspace{-2mm}
\end{figure}

\paragraph{Number of Byzantine workers.}
As increasing the number of Byzantine workers ($f_w$) does not call for increasing the total number of workers, we fix $n_w$ and hence, fixing the effective batch size in all cases.
Increasing $f_w$ leads to increasing the number of replies which the servers wait for before proceeding to the aggregation phase. Thus, a higher value of $f_w$ slightly decreases the throughput, especially in the presence of stragglers. This is confirmed in Figure~\ref{fig:throughput-f} with $f_w=3$ for running on both CPUs and GPUs. An interesting fact to note here is that increasing $f_w$ does not affect the number of iterations required for convergence since \system{} uses replies from a higher number of workers when increasing $f_w$ (e.g., $2f_w+3$).

\begin{figure}[!ht]
\centering
\vspace{-2mm}
\subfloat[CPU]{\includegraphics[width=0.49\linewidth,keepaspectratio]{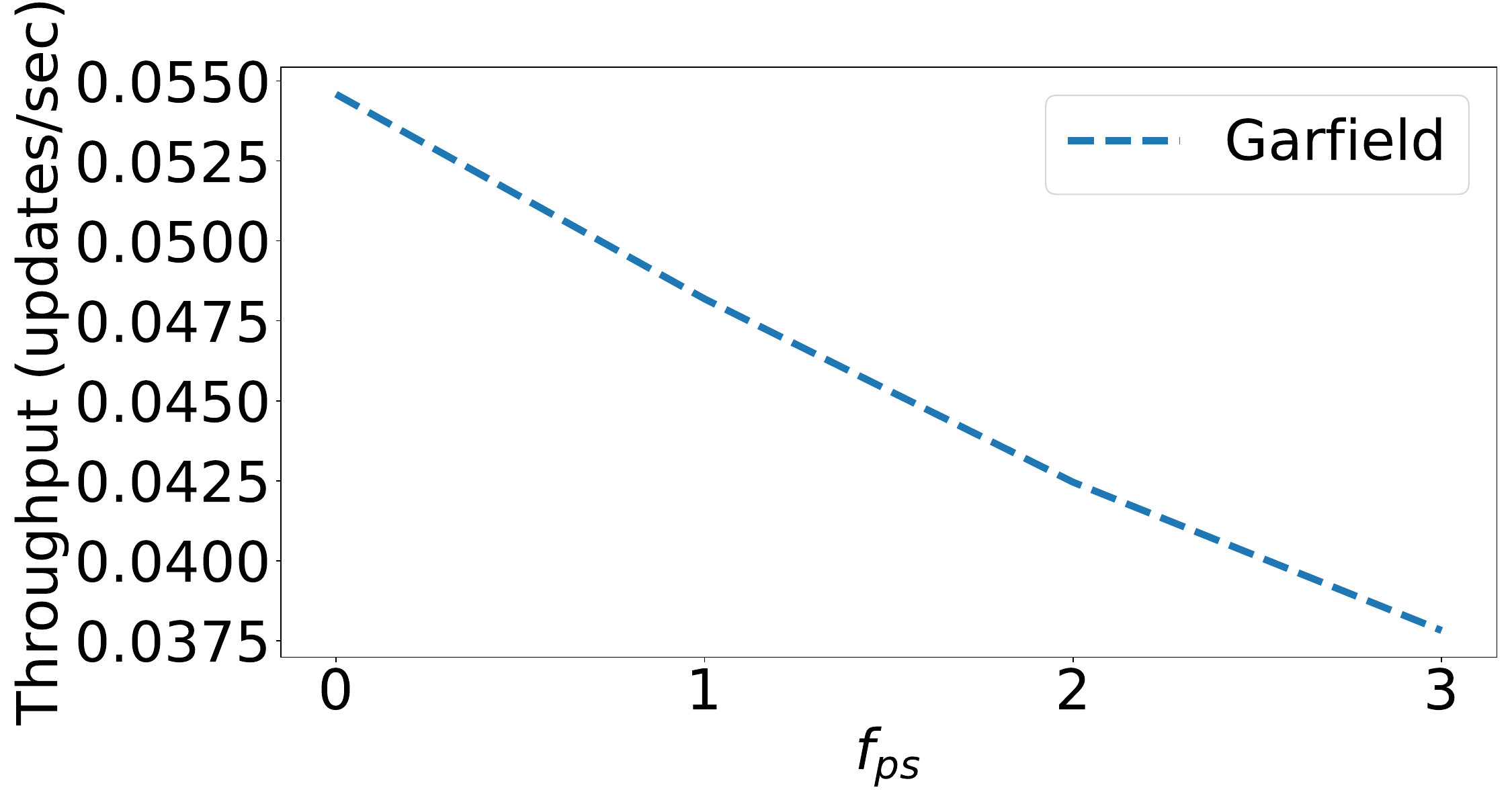}
\label{subfig:throughput-fp-tf}}
\subfloat[GPU]{\includegraphics[width=0.49\linewidth,keepaspectratio]{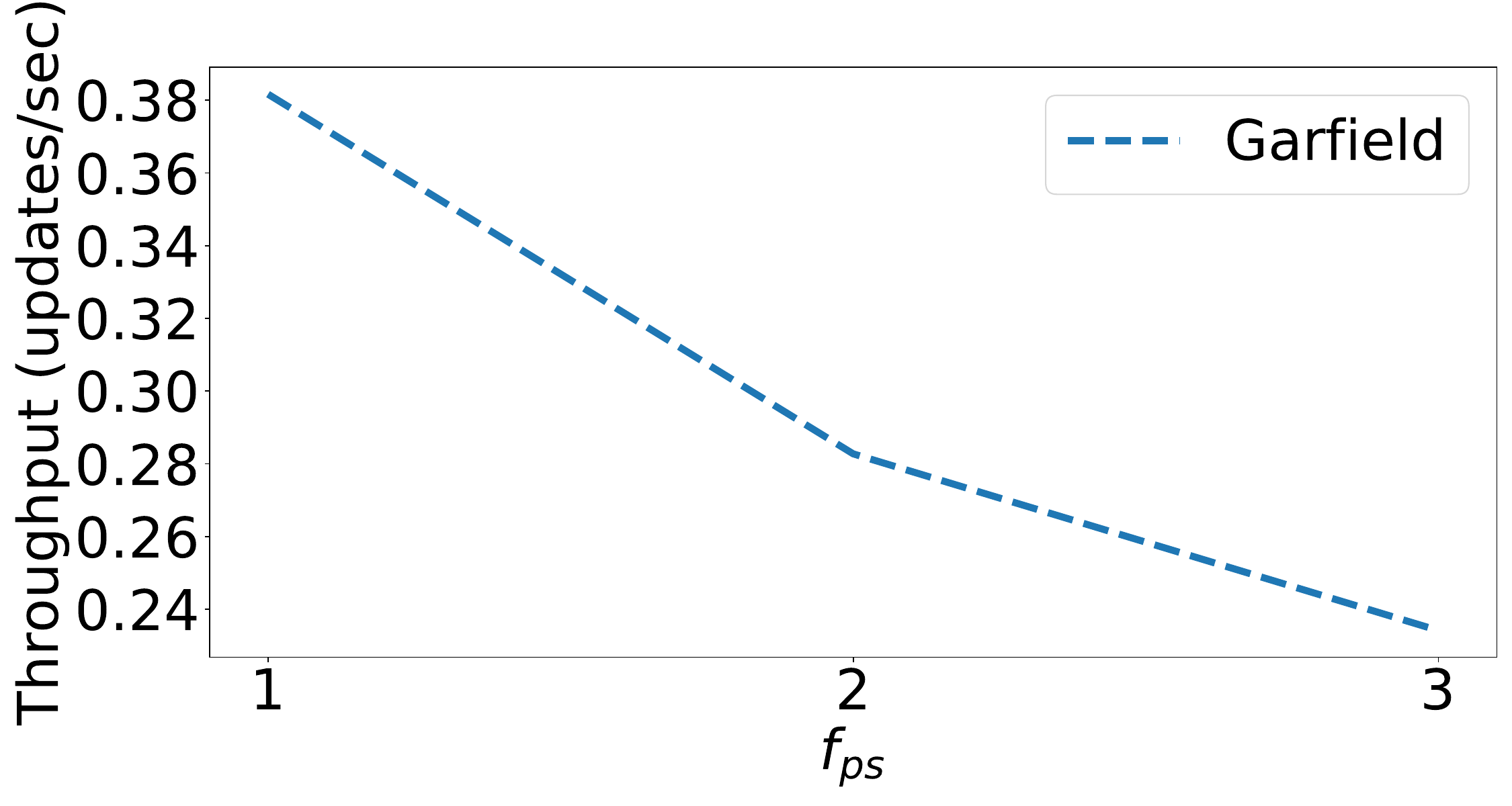}
\label{subfig:throughput-fp-pt}} 
\vspace{-2mm}
\caption{Throughput of \system{} with different number of Byzantine servers.}
\label{fig:throughput-fp}
\vspace{-2mm}
\end{figure}
\vspace{-2mm}
\paragraph{Number of Byzantine servers.}
Increasing the number of Byzantine servers ($f_{ps}$) calls for increasing the total number of the server's replicas ($n_{ps}$) so as to follow the Byzantine--resilience condition: $n_{ps} \ge 3f_{ps} + 3$ or $n_{ps} \ge 2f_{ps} + 3$ in case of using a synchronous network.
Increasing $f_{ps}$ introduces new communication links, which leads to throughput drop as shown in Figure~\ref{fig:throughput-fp}.
This throughput drop is confirmed in the \emph{state machine replication} (SMR) literature~\cite{hunt2010zookeeper,abd2005fault}, and we believe that the amount of drop (less than $45\%$ in our case) is still reasonable compared to what is reported before in the literature of SMR~\cite{cowling2006hq}. To even confirm more that this drop completely depends on the increased communication messages, we can see the same behavior (with approximately the same degradation ratio) with GPU (Figure~\ref{subfig:throughput-fp-pt}) as with CPUs (Figure~\ref{subfig:throughput-fp-tf}).

\section{Throughput in PyTorch}
\label{supp_sec:throuhgput_pytorch}
This section gives more results of \system{} implementation in PyTorch. Specifically, we show the performance of \system{} while training different models and dissect such overhead to understand its reasons.

\begin{figure}[ht]
\centering
\vspace{-2mm}
\includegraphics[width=0.49\textwidth]{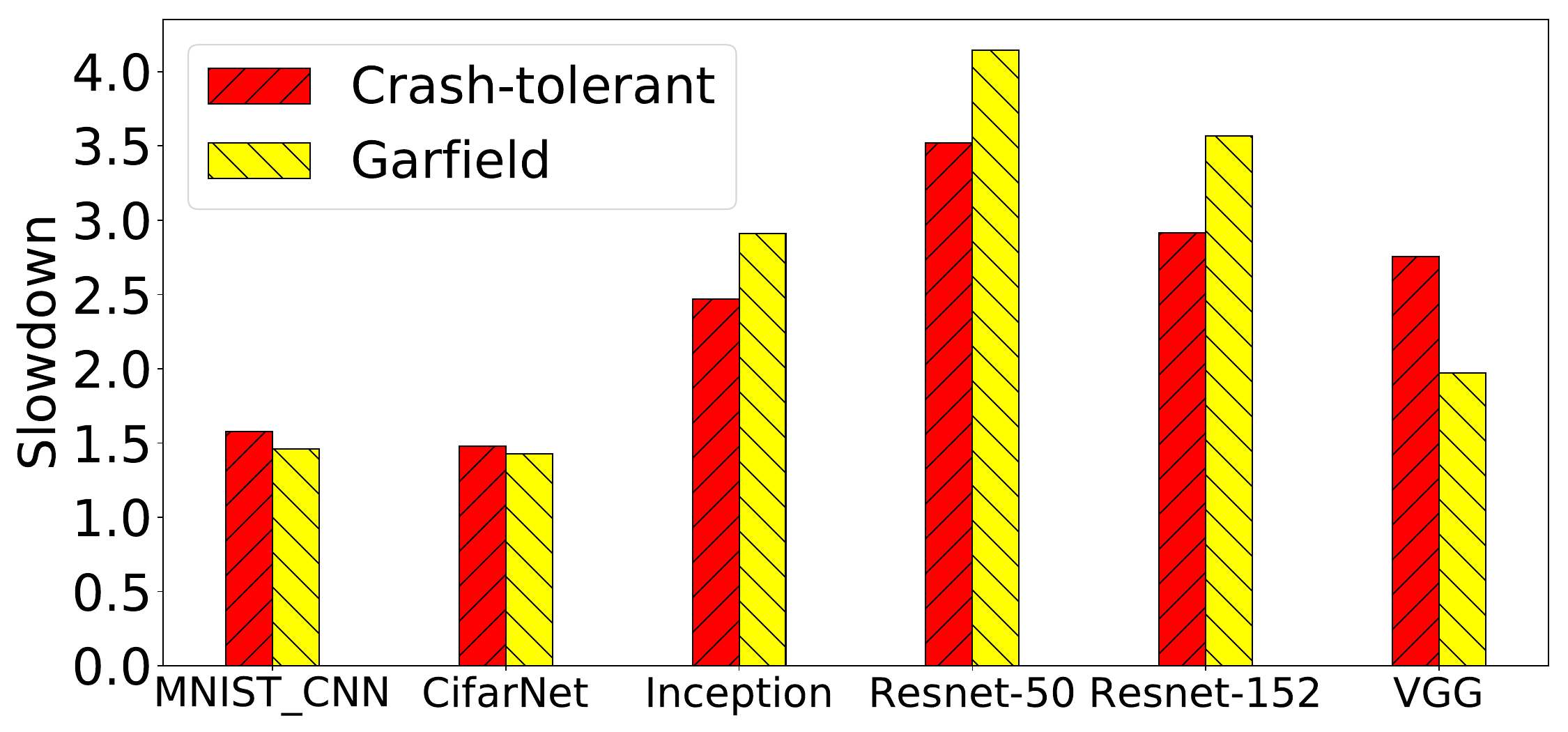}
\caption{Slowdown of fault-tolerant systems, in terms of throughput, normalized to the vanilla baseline, i.e.,\ PyTorch.}
\label{fig:models_pt}
\vspace{-2mm}
\end{figure}

\paragraph{Model dimension.}
In this section, we show the performance (basically throughput slowdown with respect to vanilla PyTorch) of our PyTorch--based implementation of \system{}, while training several models. 
We use the same models as in Figure~5 in the main manuscript (except for ResNet-200 which we replace with ResNet-152 as the former is not included in the \emph{TorchVision} models) and we use our GPU--based cluster to carry out this experiment. Figure~\ref{fig:models_pt} shows similar results to its twin (using GPUs) in the main manuscript (Figure~5b). It is emphasized in this figure that the cost of fault--tolerance is not clear with training small networks, i.e.,\ MNIST\_CNN and CifarNet. In addition, the cost of Byzantine resilience compared to crash resilience is moderate. Interestingly, our PyTorch implementation of \system{} achieves higher slowdown compared to its vanilla competitor, i.e.,\ PyTorch than what is achieved by the TensorFlow--based implementation (Figure~5b, main manuscript). 
The main reason for this slowdown is the efficiency of the communication backend that is used by PyTorch, especially when running on GPUs. Vanilla PyTorch uses the \emph{reduce()} abstraction which (1) uses GPU-to-GPU communication and (2) calculates the average of the gathered vectors immediately when received. Conversely, the \system{} communication backend uses \emph{gather()} to collect the input vectors, and the receiving end, be it a worker or a server, waits for all replies before applying the chosen GAR. The performance difference between both abstractions is emphasized more with the big models. On the TensorFlow side, both the vanilla deployment and the \system{} implementation uses gRPC so, the performance difference comes only from the difference in the number of communication rounds and the number of messages transmitted in both deployments.

\begin{figure}[h]
\centering
\vspace{-2mm}
\includegraphics[width=0.49\textwidth]{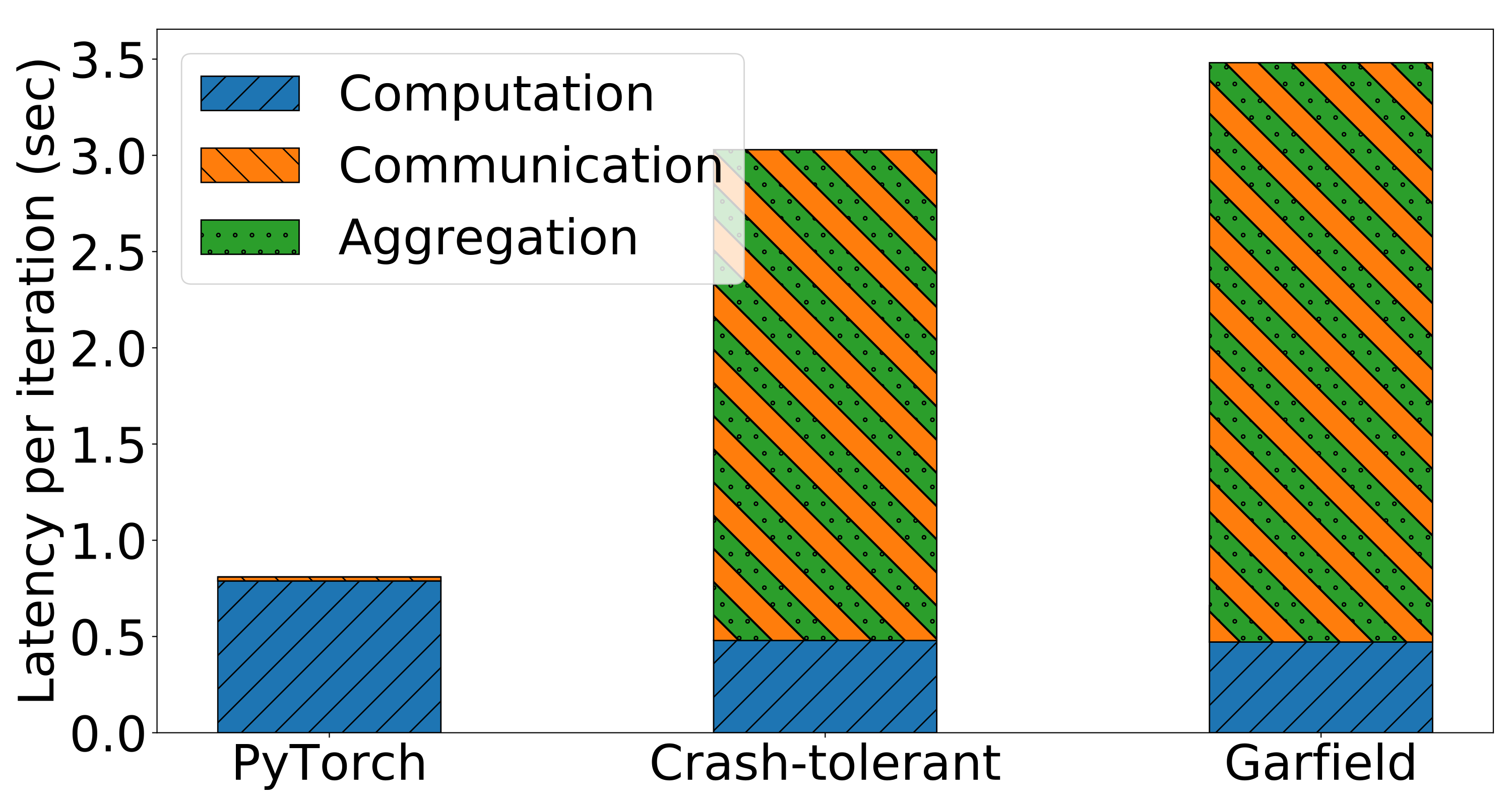}
\caption{Overhead breakdown in a PyTorch GPU-based experiment.}
\label{fig:overhead-pt}
\vspace{-2mm}
\end{figure}

\paragraph{Overhead breakdown.}
In this section, we dissect the performance of \system{} on PyTorch to understand the reasons for the overhead. Figure \ref{fig:overhead-pt} shows the time spent per iteration on computation, communication, and aggregation (if any). 
First, we can see that the fault--tolerant systems spend less time in computation than in vanilla PyTorch. This is because some of the computation done by the \system{}--enabled systems is hidden in the communication (due to our pipelined design of \system{}).
The vanilla PyTorch deployment has the lowest communication overhead compared to the fault--tolerant alternatives. This is because it uses the fast \emph{reduce()} networking abstraction which enables GPU-to-GPU communication and processes the data as they come immediately (using \emph{averaging}). 
As our design of \system{} (on PyTorch) pipelines communication with aggregation (as mentioned in Section 4.3 in the main manuscript), Figure \ref{fig:overhead-pt} shows the contribution of both in one \emph{orange--and--green} bar. Such a combined bar is higher in \system{} (the Byzantine--tolerant deployment) than the crash--tolerant deployment for two reasons. 
First, \system{} requires more communication rounds and even more messages per round compared to the crash--tolerant algorithms. Though we optimize the \emph{many-to-many} communication (e.g.,\ servers-to-workers) by parallelizing communication links and using abstractions that allow GPU-to-GPU communication, the crash--tolerant deployment can still benefit from its less demanding behavior to the network. 
Second, the crash--tolerant deployment uses the fast \emph{Average} rule for gradient aggregation and does not use any rule for aggregating models, however, \system{} uses more complex rules, e.g.,\ \mkrum{} and \brute{} to aggregate both gradients and models.

\section{Parameter Vectors Alignment}
\label{sec:param-alignment}

We do some micro--measurements to observe the alignment of the parameter vectors. 
In this section, we describe our methodology and results.

\paragraph{Methodology.} We consider the quantities $\params{i}{t}$, $\forall i \in \range{1}{n - f}$ and $t > t_s$ (our assumption must hold \emph{eventually}, e.g.,\ after some large number of steps $t_s$).
First, we calculate the differences between all parameter vectors (which we call \emph{difference vectors}) and kept ones with the $k$ highest norms.
Then, we calculate the angle between these \emph{difference vectors} to see how they are aligned.
We do that every 20 steps, throughout the training procedure, to empirically check whether the considered assumption holds or not.

\paragraph{Results.} Generally, we find that the angle between differences (from \emph{difference vectors}) of the highest norms is always close to 0\textdegree.
We give here the values of $\cos{(\phi)}$ where, $\phi$ is the angle between two difference vectors which are $a$ and $b$. This is given by $\frac{a \cdot b}{\norm{a}\norm{b}}$.
Thus, the closer this value to 1, the closer the angle to 0\textdegree.
Sample results are given in Table~\ref{table:angle}.

\begin{table}[ht]
\centering
\caption{Experimental results while studying contraction of parameter vectors.
It gives for some steps the biggest 2 norms of differences between parameter vectors collected by correct parameter servers along with $\cos{(\phi)}$ where $\phi$ is the angle between these two difference vectors. This is recorded after some large step number.}
\begin{tabular}{|c|c|c|c|c|}
\hline
Step & $\cos{(\phi)}$ & $\text{max\_diff}_1$ & $\text{max\_diff}_2$ \\ \hline
1340 & 0.9822574257850647 & 1.4122562 & 1.4163861 \\ \hline
1380 & 0.9926297664642334 & 1.3927394 & 1.3937825 \\ \hline
1400 & 0.9881128072738647 & 1.493591  & 1.5035304 \\ \hline
1440 & 0.9863847494125366 & 1.345111  & 1.3537675 \\ \hline
1480 & 0.9819352030754089 & 1.3270174 & 1.3435347 \\ \hline
\end{tabular}
\label{table:angle}
\end{table}
\end{document}